\documentclass[10pt,twocolumn,letterpaper]{article}

\usepackage{cvpr}              

\definecolor{cvprblue}{rgb}{0.21,0.49,0.74}
\usepackage[pagebackref,breaklinks,colorlinks,allcolors=cvprblue]{hyperref}

\usepackage[capitalize]{cleveref}
\crefname{section}{Sec.}{Secs.}
\crefname{table}{Tab.}{Tabs.}
\crefname{equation}{Eq.}{Eqs.}
\usepackage{comment}
\usepackage{multirow}
\usepackage{graphicx}
\usepackage{makecell}  
\usepackage{pifont}
\usepackage{tcolorbox}
\usepackage[normalem]{ulem}
\usepackage{soul}
\newcommand{\customfootnotetext}[2]{{
\renewcommand{\thefootnote}{#1}
\footnotetext[0]{#2}}}
\definecolor{lightblue}{rgb}{0.8,0.9,1}
\definecolor{lightyellow}{rgb}{1,1,0.8}
\definecolor{lightpurple}{rgb}{0.9,0.8,1}
\definecolor{peach}{rgb}{1,0.9,0.8}
\definecolor{mint}{rgb}{0.8,1,0.9}
\sethlcolor{peach}

\newcommand{\hlyellow}[1]{\sethlcolor{lightyellow}\hl{#1}}
\newcommand{\hlpurple}[1]{\sethlcolor{lightpurple}\hl{#1}}
\newcommand{\hlblue}[1]{\sethlcolor{lightblue}\hl{#1}}
\newcommand{\hlgreen}[1]{\sethlcolor{mint}\hl{#1}}

\DeclareRobustCommand{\erase}{\bgroup\markoverwith{\textcolor{red}{\rule[.5ex]{2pt}{0.4pt}}}\ULon}

\newcommand{\cmark}{\ding{51}} 
\newcommand{\xmark}{\ding{55}} 

\definecolor{maskgreen}{rgb}{0.639,0.984,0.639}
\definecolor{maskpink}{rgb}{1,0.803,0.973}
\definecolor{maskorange}{rgb}{0.988,0.827,0.635}
\definecolor{maskblue}{rgb}{0.624,0.804,1}
\definecolor{nvgreen}{rgb}{0.462,0.725,0}
\newcommand{\hlmaskpink}[1]{\sethlcolor{maskpink}\hl{#1}}

\newcommand{\hlmaskorange}[1]{\sethlcolor{maskorange}\hl{#1}}
\newcommand{\hlmaskblue}[1]{\sethlcolor{maskblue}\hl{#1}}

\newcommand{\methodabbr}{Omni-RGPT}
\newcommand{\featuremarkabbr}{Token Mark}
\newcommand{\datasetabbr}{RegVID-300k}

\def\paperID{} 
\def\confName{CVPR}
\def\confYear{2025}

\title{Omni-RGPT: Unifying Image and Video Region-level Understanding \\ via Token Marks}

\author{Miran Heo\textsuperscript{1,2,$\dagger$}, Min-Hung Chen$^1$, De-An Huang$^1$, Sifei Liu$^1$, Subhashree Radhakrishnan$^1$\\Seon Joo Kim$^{2}$, Yu-Chiang Frank Wang$^{1, 3}$, Ryo Hachiuma$^1$
\\
$\textsuperscript{\rm 1}$NVIDIA,\quad$\textsuperscript{\rm 2}$Yonsei University,\quad$\textsuperscript{\rm 3}$National Taiwan University\\
\href{https://miranheo.github.io/omni-rgpt}{\texttt{\textcolor{magenta}{https://miranheo.github.io/omni-rgpt}}}
}

\begin{document}
\twocolumn[{%
\renewcommand\twocolumn[1][]{#1}%
\maketitle
\centering
\vspace{-7mm}
\includegraphics[width=0.95\linewidth]{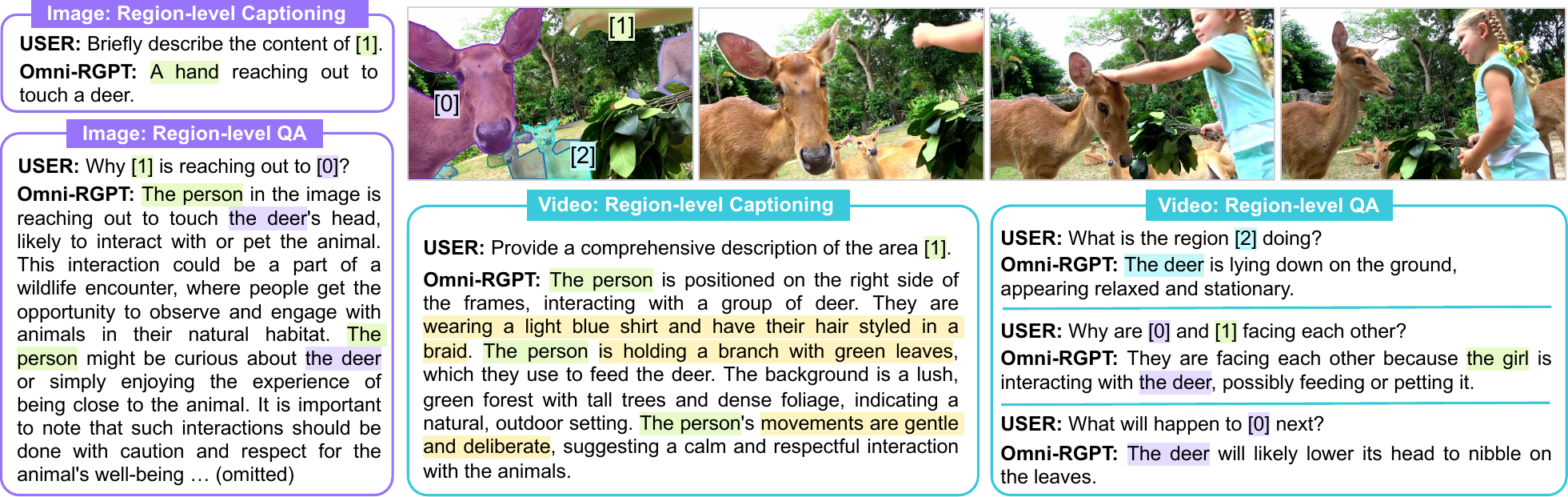}
\captionof{figure}{\textbf{Representative demo examples of \methodabbr{}.} We introduce a unified multimodal large language model that integrates region-level understanding for both images and videos. Given user-defined localized region inputs (boxes or masks) accompanied by a corresponding text prompt, \methodabbr{} generates responses tailored to the visual context of each region for both images and videos.\protect\footnotemark}
\label{fig:teaser_main}
\vspace{1em} 
}]

\customfootnotetext{}{$\dagger$ Miran Heo was an intern at NVIDIA during the project. }

\begin{abstract}
We present \methodabbr, a multimodal large language model designed to facilitate region-level comprehension for both images and videos.
To achieve consistent region representation across spatio-temporal dimensions, we introduce \featuremarkabbr, a set of tokens highlighting the target regions within the visual feature space.
These tokens are directly embedded into spatial regions using region prompts (e.g., boxes or masks) and simultaneously incorporated into the text prompt to specify the target, establishing a direct connection between visual and text tokens.
To further support robust video understanding without requiring tracklets, we introduce an auxiliary task that guides \featuremarkabbr{} by leveraging the consistency of the tokens, enabling stable region interpretation across the video.
Additionally, we introduce a large-scale region-level video instruction dataset (\datasetabbr).
\methodabbr{} achieves state-of-the-art results on image and video-based commonsense reasoning benchmarks while showing strong performance in captioning and referring expression comprehension tasks.
\end{abstract}    
\section{Introduction}
\label{sec:intro}
Multimodal large language models (MLLMs) have evolved to interpret visual elements, progressing from text prompts for holistic image understanding to sophisticated approaches for region-level understanding. To achieve interactive region-specific comprehension in images, recent methods employ various strategies to represent target regions: encoding textual box coordinates within the text tokens~\cite{ChatSpot, VistaLLM, ChatterBox, Shikra, kosmos2, MiniGPT4, AllSeeingv2, LION}, utilizing visual RoI features~\cite{RGPT, Groma, GPT4RoI, AllSeeing, Osprey, AnyRef}, or applying visual markers~\cite{SoM, ViP-LLaVA}. Extending these capabilities to the video domain, some approaches~\cite{Elysium, Merlin} incorporate initial frame bounding box coordinates as a textual form for the region-level video understanding tasks.
Nonetheless, a general approach that effectively addresses region-specific tasks across both image and video remains an open challenge.\footnotetext{Our model refers to the girl in the video as ``they'' as a singular term to alleviate ethical issues instead of ``she'', similar to ChatGPT~\cite{GPT4o}.}

\begin{figure}[t]
\begin{center}
\includegraphics[width=\linewidth]{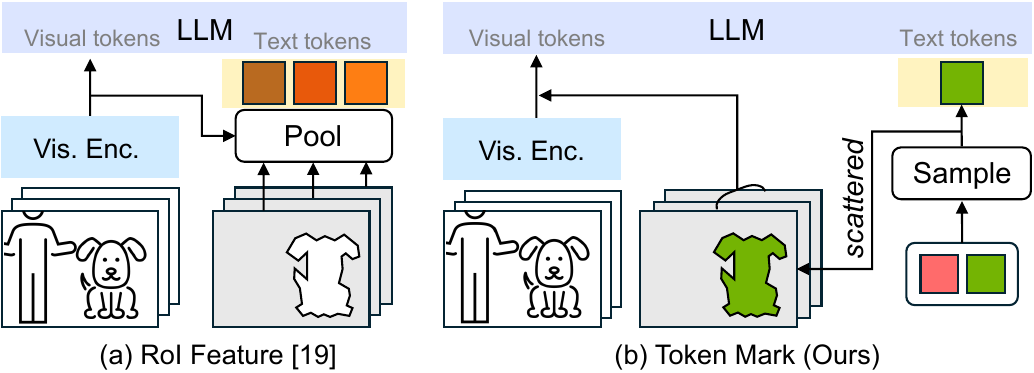}
\end{center}
\vspace{-5mm}
\caption{
\textbf{Method comparison.} (a) RoI-based methods generate visual region prompts using RoI-aligned visual features, potentially leading to temporal drift in the visual features of the target object in the video domain. (b) In contrast, our \featuremarkabbr{} is assigned to the corresponding region, preserving a consistent spatio-temporal target reference.
}
\label{fig:teaser_method}
\vspace{-5mm}
\end{figure}

One key challenge in developing a solution is achieving scalability for video sequences. Since videos can contain a large number of frames, approaches that rely on bounding box coordinates as textual input face scaling limitations, as input region tokens increase linearly with the number of frames.
RoI-based methods~\cite{RGPT, GPT4RoI} also encounter this issue, as they require repeated extraction of visual features from spatial regions (see \cref{fig:teaser_method} (a)).
Relying on a single frame (\eg, the initial frame) as an alternative~\cite{Elysium, Merlin} is also suboptimal, as it lacks a robust reference for the target across subsequent frames.

Another challenge is addressing the temporal drift issue.
There is no standardized method for unifying the multiple vectors representing the same object across different frames (\eg, bounding boxes in each frame) into a single, consistent vector. Unlike in static images, this issue becomes particularly problematic in videos, as target objects often change in appearance across frames due to motion, scale shifts, and perspective changes. Consequently, merging RoI features into a single representation can introduce inconsistencies, resulting in a loss of essential visual details.
We identify that a key limitation of previous methods~\cite{RGPT, GPT4RoI} is the reliance on representations that may not consistently capture regions across frames, particularly when aiming for a unified solution for both images and videos.

In this paper, we present \textbf{\textit{\methodabbr}}, a region-level MLLM designed for both images and videos. At the core of our framework is \textit{\featuremarkabbr}, a novel region representation that enables seamless region-level understanding across both inputs (\cref{fig:teaser_method} (b)).
Our key insight is to \textit{invert} the traditional approach: rather than generating region embeddings from visual features, we predefine a set of tokens, using them as markers to identify regions within the latent space.
Given visual-text inputs paired with target region prompts (\eg, boxes or masks), we sample a token mark and embed it within the spatial location defined by the region prompt. This embedding is further injected into the corresponding text prompt, allowing the LLM to directly reason the alignment between visual regions and text prompts. 
This approach effectively addresses two key challenges: 1) \textit{Scalability}—since each target has a unique representation shared across frames, the number of input text tokens remains independent of the number of frames, and 2) \textit{Temporal drift}—representing each target as a token ensures consistent reference across frames.

Building on \featuremarkabbr{}, we introduce \textit{Temporal Region Guide Head}, an auxiliary head specifically designed for video input to address the limitations of tracking-dependent region prompts (\ie, tracklets), which are often impractical in real-world applications.
Using the region prompt from the initial frame, this auxiliary head operates on the LLM's output visual tokens, classifying each visual token according to its assigned token marks.
The representation of \featuremarkabbr{} supports effective region guidance during training, enabling robust and consistent region understanding across frames during inference without the need for full tracklets and additional cost (see \cref{fig:teaser_main}).

Although the capabilities of MLLMs are heavily dependent on large-scale data~\cite{LLaVA, VideoChatGPT}, the research community lacks video instruction data for region-level understanding. Therefore, we introduce a large-scale, diverse, and fine-grained \textbf{Reg}ion-level \textbf{Vi}deo \textbf{I}nstruction \textbf{D}ataset (\textbf{\datasetabbr}). The dataset includes 98k unique videos, with 214k regions curated from 10 public video datasets and 294k region-level instruction samples. We present an automated pipeline for curating the large-scale region-level video instruction samples based on powerful GPT4o~\cite{GPT4o}.

We demonstrate the capabilities of \methodabbr{} across both image and video inputs on diverse region-specific comprehension tasks, including visual commonsense reasoning, captioning, and referring expression comprehension. Notably, we achieve state-of-the-art performance on challenging commonsense reasoning benchmarks, excelling in image-based (VCR~\cite{VCR}) and video-based (Causal-VidQA~\cite{Causal-VidQA}) tasks. 
Supported by our newly introduced \datasetabbr{} dataset, \methodabbr{} also exhibits enhanced capabilities for video captioning. Additionally, it performs strongly in region-level understanding tasks in images, such as captioning and referring expression comprehension.

\begin{figure*}[t]
\begin{center}
\includegraphics[width=\linewidth]{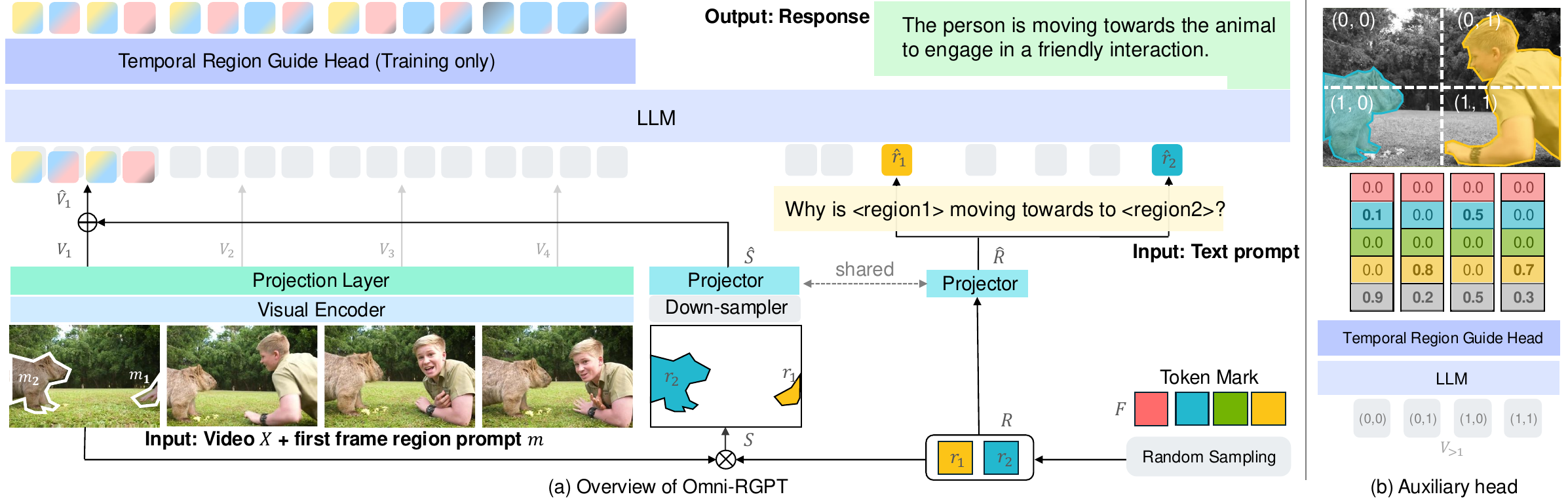}
\end{center}
\vspace{-5.4mm}
\caption{
\textbf{(a) Overview:} \methodabbr{} enables region-level understanding across image and video inputs. Given region prompts (\eg boxes or masks) in a single image or the initial frame of a video, we assign \textit{\featuremarkabbr{}} --- a set of vectors serving as spatio-temporal region indicators --- to the region.
These vectors are embedded into the spatial region localized by the region prompt and directly injected into both visual and text prompts to indicate the target. \textbf{(b) Auxiliary Head:} We further introduce \textit{Temporal Region Guide Head} to achieve robust region understanding in videos without relying on tracklet prompts. Building on Token Mark's consistent representation of target objects across frames, this auxiliary task classifies the target Token Mark for visual tokens in subsequent frames.
}
\label{fig:main}
\vspace{-3.3mm}
\end{figure*}

\section{Related Work}
\label{sec:related_work}
\subsection{Vision-Language MLLMs}
Recent advancements in vision-language MLLMs such as GPT-4V~\cite{GPT-4V} and Gemini~\cite{gemini}, along with open-source counterparts like LLaVA~\cite{LLaVA}, VILA~\cite{VILA}, and QwenVL~\cite{QwenVL}, have demonstrated robust capabilities in interpreting visual inputs. 
These models leverage multimodal capabilities to bridge visual and linguistic information, enhancing performance on tasks requiring visual comprehension.
Building on these advancements, recent efforts have aimed to extend these models to video understanding~\cite{VideoChatGPT, VideoChat, Video-LLaMA, InternVideo, LLaVA-Next-Video, Slowfast-LLaVA, GroundingGPT}. 
However, while current approaches address global visual comprehension, achieving fine-grained, region-specific understanding in both images and videos remains an open challenge.
This limitation is particularly evident in interactive settings, where accurate interpretation of localized visual details is essential.

\subsection{Region-level MLLMs}
\noindent \textbf{Image-based.}
Early methods employ numerical bounding box coordinates as textual prompts to specify visual regions within the LLM’s input~\cite{ChatSpot, VistaLLM, ChatterBox, Shikra, kosmos2, MiniGPT4, AllSeeingv2, LION, collavo}.
By encoding region-level cues as text, these approaches facilitate seamless integration with global vision-language MLLMs.
Instead of directly using bounding box coordinates as text, KOSMOS-2~\cite{kosmos2} introduces location tokens, which embed these coordinates as discretized representations.

Another line of work leverages RoI-aligned visual features derived from region prompts (\textit{e.g.}, bounding boxes or masks)~\cite{ASM, Groma, GPT4RoI, RGPT}.
ASM~\cite{ASM} introduces a location-aware image tokenizer, a transformer decoder module that uses RoI features as query inputs to generate region-level tokens for LLMs.
Similarly, VisionLLM~\cite{VisionLLM} and RegionBLIP~\cite{RegionBLIP} apply transformer decoders to derive region-specific tokens.
In contrast, GPT4RoI~\cite{GPT4RoI} and RegionGPT~\cite{RGPT} directly embed extracted RoI features into region-assigned tokens within language prompts.

An alternative approach employs visual markers~\cite{SoM, ViP-LLaVA}. 
SoM~\cite{SoM} offers a zero-shot solution using GPT-4V by overlaying masks with numbered markers on the image, which GPT-4V’s OCR capabilities interpret.
While effective, this method may alter the image’s original appearance, potentially affecting details like color, and relies on rule-based algorithms to determine marker size and placement.
In contrast, ViP-LLaVA~\cite{ViP-LLaVA} enables more interactive, user-defined prompts (\eg, free-form scribbles), allowing for a flexible and intuitive approach to region understanding.


\noindent \textbf{Video-based.}
In contrast to the image domain, region-aware comprehension in the video domain remains relatively underexplored.
Elysium~\cite{Elysium} and Merlin~\cite{Merlin} input bounding box coordinates from the initial frame as text, enabling object tracking by predicting bounding box coordinates across the sequence.
Additionally, Elysium introduces a new dataset; however, it provides only brief descriptions of dominant objects with limited temporal detail.
MotionEpic~\cite{VoT} takes a different approach by using RoI information as nodes within a dynamic scene graph representation for videos.
Initialized with frame-level proposals, they build a spatio-temporal scene graph through recurrent encoding.
Unlike the above approaches, we present a novel approach that flexibly supports both box and mask representations for videos without relying on off-the-shelf tracking methods.
Additionally, we introduce the instruction dataset to enhance region-level understanding capability.
\section{\methodabbr}
\label{sec:method}

We visualize our architecture in \cref{fig:main}.
We build on the core design of LLaVA~\cite{LLaVA}, where the input image or video $ X \in \mathbb{R}^{T \times 3 \times H_0 \times W_0}$ (with $ T=1 $ for images) is processed by a vision encoder $ f( \cdot ) $, producing visual features.
Through a projection layer, these visual features are then projected into visual tokens $ V \in \mathbb{R}^{T \times D \times H \times W}$, where $D$ is the input dimension of LLM.
The visual tokens are then processed by the LLM $\mathcal{F}_{\text{LLM}}(\cdot )$ with a text prompt, which enables joint reasoning across textual and visual modalities.

Our objective is to enable the model to understand specific visual elements in response to an input text prompt by incorporating $N$ input region prompts $\{m_i\}_{i=1}^N$, where each $m_i \in \{ 0, 1 \}^{H_0 \times W_0}$ defines a target region (\eg, bounding box or mask).
These region prompts, corresponding to a special token \texttt{<region>} as a placeholder in the text prompt, serve to identify and infer designated areas across the spatio-temporal dimension.

At a high-level, we have a set of tokens (\featuremarkabbr{}), which can be thought of as different paint colors on a palette.
We randomly select a color to represent each target specified by the region prompt.
In \cref{fig:main} (a), two tokens are chosen to represent the ``person'' and ``animal'', respectively.
This color is then applied to both visual and text token prompts. For visual tokens, we create a blank canvas and apply the selected color to the specified regions, overlaying this colored canvas onto the visual tokens. For text tokens, we replace the target placeholder (\eg, \texttt{<region>}) with its assigned token.
Through this process, the model learns ``where to look'' during training by internalizing the predefined palette.
In \cref{sec:feature_mark}, we provide the concrete formulation of \featuremarkabbr{}, and introduce an auxiliary head (\cref{fig:main} (b)) specifically tailored for video input in \cref{sec:aux_classifier}.

\subsection{\featuremarkabbr{}}
\label{sec:feature_mark}
We define \featuremarkabbr{} as a set of tokens $ F \in \mathbb{R}^{N_F \times C}$, where $N_F$ is the total number of tokens and $C$ denotes the feature dimension.
To represent a region using \featuremarkabbr{}, we uniformly sample $N$ indices from $[N_F]$ without replacement, obtaining the set of tokens $ R = \{r_i\}_{i=1}^N$.
Each sampled token $r_i$ is then matched one-to-one with corresponding region prompt $m_i$ so that the $i$-th \featuremarkabbr{} aligns with the $i$-th region prompt.
These tokens serve as spatio-temporal region indicators and are injected into the language-side input for the associated visual content.
Specifically, we project the \featuremarkabbr{} directly into the word embedding space using a linear layer: $\hat{R} = \mathcal{F}_{\text{proj}}(R) \in \mathbb{R}^{N \times D}$.

To associate the sampled \featuremarkabbr{} $r_i$ with its corresponding region $m_i$, we embed the tokens into the relevant pixels defined by the region prompts.
Specifically the Spatial \featuremarkabbr{} $S \in \mathbb{R}^{C \times H_0 \times W_0}$ at each pixel location $(h, w)$ is computed as:
\begin{equation}
    S_{:, h, w} = \frac{\sum_{i=1}^{N} m_{i, h, w} \cdot r_i}{\epsilon + \sum_{i=1}^{N} m_{i, h, w}},
    \label{eq:spatial_feature_mark}
\end{equation}
where $\epsilon$ is a small positive constant added to prevent division by zero when no masks are active at position $(h, w)$.

Next, we then downscale $ S $ to match the shape of the visual tokens $V$ by applying adaptive average pooling, resulting in the updated Spatial \featuremarkabbr{} $ \tilde{S} $.
We then project it into the same feature space as $\hat{R}$ using the shared projection layer, yielding $\hat{S} = \mathcal{F}_{\text{proj}}(\tilde{S}) \in \mathbb{R}^{D \times H \times W}$.
Finally, we integrate the spatial region-specific information into the visual tokens: $\hat{V} = V + \hat{S}$.

The proposed approach offers several key advantages. \textbf{i) Preventing temporal drift:} By encoding the target region as unique representation shared across frames, our method ensures consistent region assignments throughout video sequences.
This consistency distinguishes our approach from RoI-based methods, where representations of target objects often vary across frames.
\textbf{ii) Direct region-language connection:} Projecting \featuremarkabbr{} directly within the word embedding space enables efficient modeling of region-language relationships.
Unlike methods that rely on textual descriptions for each region, our approach facilitates seamless user interaction without additional textual input for the region.
\textbf{iii) Preserving vision-language global alignment:} By incorporating region information as residual features, our architecture retains alignment with the base image-text pair multimodal framework (\eg, LLaVA). 
In cases without region prompts, the model functions identically to the base architecture.

\begin{figure*}[t]
\begin{center}
\includegraphics[width=\linewidth]{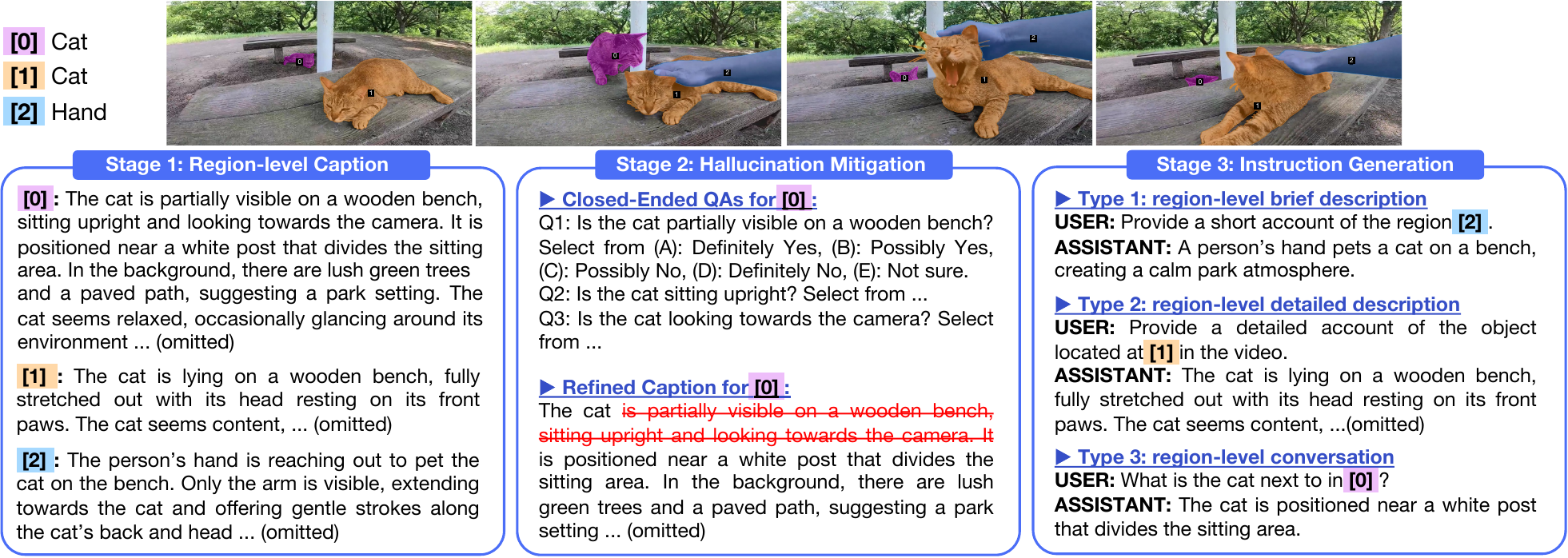}
\end{center}
\vspace{-4mm}
\caption{
\textbf{Overview of our instruction sample generation pipeline.} From a video with region masklets and nouns, the region-level captions, which contain contextual and temporal information about regions, are generated from GPT4o (left). Then, the hallucinations in the captions are mitigated (middle). Lastly, the instruction samples that cover diverse aspects of the regions are generated (right). 
}
\label{fig:data_visualization}
\end{figure*}

\subsection{Temporal Region Guide Head}
\label{sec:aux_classifier}
For video input, an auxiliary head (\cref{fig:main} (b)) is introduced during training to enhance region consistency across frames, ensuring an accurate representation of regions even when a region prompt is provided for only the first frame.
This auxiliary head classifies the corresponding \featuremarkabbr{} for each visual token, implicitly guiding the model to understand the target region without relying on explicit video object correspondence from tracklets.

Let $V_t$ represent the visual tokens at the $t$-th frame, forming a sequence of visual tokens for the entire video, denoted as $V_{vid} = (\hat{V}_1, V_2, \dots, V_T)$, where $\hat{V}_1$ contains the target region information. 
The sequence $V_{vid}$ is then processed by the language model, which aims to generate region-aware predictions for the entire video sequence.

The auxiliary classification head $\mathcal{F}_{\text{aux}}$ performs as:
\begin{equation}
    \mathcal{F}_{\text{aux}}(\mathcal{F}_{\text{LLM}}(V_{vid})) \in \mathbb{R}^{T \times H \times W \times (N_F+1)},
    \label{eq:aux_cls}
\end{equation}
where $N_F+1$ is classification categories (the $N_F$ \featuremarkabbr{} and the background).

Since the visual tokens are downscaled from the original input resolution, multiple \featuremarkabbr{}s may exist within a single visual token. 
To handle this, we apply soft-label classification, assigning each token a soft-label distribution over the $N_F+1$ categories to reflect the proportion of each token belonging to multiple regions or the background.

\noindent \textbf{Loss Function.} 
The final loss is defined as $\mathcal{L} = \mathcal{L}_{\text{LLM}} + \alpha \mathcal{L}_{\text{aux}}$, where $\alpha$ balances the contribution of the auxiliary classification loss.
The language model loss, $\mathcal{L}_{\text{LLM}}$, is computed as the cross-entropy loss between the predicted tokens and the ground truth tokens.
 Meanwhile, the auxiliary classification loss, $\mathcal{L}_{\text{aux}}$, is defined as the cross-entropy loss between the predicted soft-label distributions and the ground truth soft-label distributions for each visual token.
This region guide head is used only during training and does not introduce additional latency during inference.
\section{Region-level Video Instruction Dataset}
\label{sec:dataset}

\begin{table}[tb]
\centering
\resizebox{\columnwidth}{!}{
\begin{tabular}{l|ccccc}
\toprule
\multirow{2}{*}{Datasets} & \multirow{2}{*}{Videos} &  \multirow{2}{*}{Regions}  & \multirow{2}{*}{\begin{tabular}{c}Caption  \\ Length\end{tabular}} & \multirow{2}{*}{\begin{tabular}{c}Video \\ Sources\end{tabular}} & \multirow{2}{*}{\begin{tabular}{c}Instruction \\ Samples\end{tabular}} \\ 
& & & & & \\ \midrule
Vid-STG~\cite{VidSTG} & 7.8k & 44k & 10.8 & 1 & - \\
HC-STVG~\cite{HCSTVG} & 12k & 12k & 17.3 & 1 & - \\
LV-VIS~\cite{LVVIS} & 4.8k & 26k & 1 & 1 & - \\
BenSMOT~\cite{BenSMOT} & 3.3k & 7.7k & \underline{39.9} & 1 & - \\ 
Elysium~\cite{Elysium} & \textbf{1.2M} & \textbf{1.2M} & 2.2 & 1 & - \\ \midrule
V-ChatGPT~\cite{VideoChatGPT} & 13k & - & - & 1 & 100k \\
Valley~\cite{Valley} & 65k & -  & - & 1 & 65k \\
VCG~\cite{VideoGPT+} & 6.2k & - & - & 1 & \underline{112k} \\ \midrule \midrule
Ours  & \underline{98k} & \underline{214k} & \textbf{59.7} & \textbf{10} & \textbf{294k}  \\ \bottomrule
\end{tabular}}
\caption{\textbf{Comparison of our \datasetabbr{} with existing video datasets.} Our dataset is the first region-level video instruction dataset that contains diverse video sources with detailed captions.}
\label{tab:dataset_comparison}
\vspace{-2mm}
\end{table}

\noindent \textbf{Overview.} Since there is no existing region-level video instruction dataset, we propose \textbf{\datasetabbr{}} to enhance MLLMs' dialog capability and obtain accurate responses about the regions in the videos (\cref{fig:data_visualization}). Inspired by GPT4V-assisted detailed captioning~\cite{sharegpt4v, sharegpt4video, Groma} and caption-guided instruction dataset~\cite{LLaVA, Groma, VideoChatGPT}, our approach consists of three-steps, i) GPT4o-assisted region-level detailed captioning, ii) visual hallucination mitigation, and iii) caption-guided region-level instruction sample generation. Detailed information (\eg, full video sources and GPT4o text prompts) are shown in the supplementary material.

\noindent \textbf{Key Characteristics.} 
Our dataset statistics and comparison with existing region-level video caption~\cite{VidSTG, LVVIS, HCSTVG, BenSMOT,Elysium} and common video instruction datasets~\cite{VideoChatGPT, Valley, VideoGPT+} are summarized in \cref{tab:dataset_comparison}. For example, although Elysium~\cite{Elysium} contains 1.2M videos, the region captions are only a single or few words (\eg, nouns, phrases), and the videos are collected only from WebVid~\cite{WebVid}. The key characteristics of our dataset are \textbf{i) large-scale}: our dataset consists of 98k unique videos, 214k tracklets or masklets, and 294k instructions, such as region-level detailed captioning, conversations, \textbf{ii) diverse}: the videos are collected from 10 public datasets used in different tasks, \textbf{iii) fine-grained QAs}: each region is described within about 60 words, including contextual and temporal information of the regions, resulting in diverse instruction samples, and \textbf{iv) high-fidelity}: the visual hallucinations in detailed captions are mitigated.

\noindent \textbf{Data Collection.} 
The videos are collected from 10 public datasets~\cite{MeViS,RefYTVOS,LVVIS,VidSTG,HCSTVG,ANetEntities,AVAKinetics,ActionGenome,BDD100k,Elysium} that contain annotated regions (\eg, masklets, tracklets, or a single frame bounding box) along with nouns.

\subsection{GPT4o-Assisted Region-level Captions}
From paired videos and masklets of regions\footnote{For datasets with only tracklets or a single bounding box, we apply SAM~\cite{SAM} or SAM2~\cite{SAM2} to generate masklets.}, we adapt the visual prompting technique of SoM~\cite{SoM} to overlay object masks with region indices at the center of each mask for every frame in the video. We then input the SoM-processed videos into GPT4o~\cite{GPT4o}, requesting enriched captions by including contextual and temporal information of each masklet from nouns in text prompts, such as ``Generate the detailed description of [0]: cat, [1]: cat, [2]: hand''.

\subsection{Visual Hallucination Mitigation}
We mitigate the visual hallucination in the generated captions to improve the fidelity. Although the region-level captions generated by GPT4o contain fine-grained information, the synthetically generated detailed captions contain visual hallucinations~\cite{UnveilDetCap, VisualHallucinations}, and it is crucial to mitigate these to generate high-fidelity instruction samples.

\begin{table*}[t]
\centering
\small
\begin{tabular}{lcc|c|c|ccc|ccc|c}
\toprule
\multirow{2}{*}{\makecell{Models}}  &\multirow{2}{*}{LLM size} & Category &\multirow{2}{*}{\makecell{Acc@D}}  & \multirow{2}{*}{\makecell{Acc@E}} & \multicolumn{3}{c|}{Acc@P} & \multicolumn{3}{c|}{Acc@C}  &\multirow{2}{*}{\makecell{Acc@All}}\\
                         &     &  input  & &       & A   & R   & AR  & A   & R   & AR  \\
\midrule
VGT$^\dag$~\cite{VGT}                & -  & \checkmark & 70.8 & 70.3 & 55.2 & 56.9 & 38.4 & 61.0 & 59.3 & 42.0  & 55.4\\
TranSTR~\cite{TranSTR}                & - & \checkmark & 73.6 & 75.8 & 65.1 & 65.0 & 48.9 & 68.6 & 65.3 & 50.3 & 62.2  \\ \midrule
Video-LLaMA$^\ddag$~\cite{Video-LLaMA}   & 7B  & \checkmark & 69.2 & 71.0 & 63.6 & 62.4 & 44.4 & 65.4 & 60.1 & 45.0 & 57.4 \\
VideoChat$^\ddag$~\cite{VideoChat}       & 7B  & \checkmark & 72.9 & 73.9 & 65.2 & 63.1 & 45.9 & 66.0 & 62.7 & 45.8 & 59.6\\
Video-ChatGPT$^\ddag$~\cite{VideoChatGPT} & 7B & \checkmark & 73.1 & 75.1 & 66.0 & 63.9 & 46.0 & 67.8 & 63.6 & 50.0 & 61.1\\
Video-LLaVA$^\ddag$~\cite{Video-LLaVA}    & 7B & \checkmark & 73.7 & 74.4 & 67.6 & 65.4 & 47.7 & 68.0 & 64.9 & 51.5 & 61.8\\
MotionEpic~\cite{VoT}                   & 7B & \checkmark & \underline{81.2} & \underline{83.0} & \underline{74.3} & \underline{73.7} & \underline{54.7} 
                                        & \underline{74.5} & \underline{73.8} & \underline{58.6}  & \underline{69.4} \\
\midrule
\methodabbr{}                                   &7B & & \textbf{84.0} & \textbf{84.6} & \textbf{84.2} & \textbf{85.4} & \textbf{76.9} 
                                        & \textbf{74.7} & \textbf{74.0} & \textbf{64.3}  & \textbf{77.5}\\
\methodabbr{}                                   &13B & & 84.5 & 85.1 & 85.5 & 85.6 & 78.0 
                                        & 76.7 & 75.3 & 67.5  & 78.8\\
\bottomrule
\end{tabular}
\vspace{-2mm}
\caption{\textbf{Accuracy (\%) on Causal-VidQA validation set}, it consists of four sub-tasks: Description (D), Explanation (E), Prediction (P), and Counterfactual Reasoning (C), each presented as multiple-choice questions with an answer (A) component.
Prediction and Counterfactual tasks additionally include a reason (R) choice. $\dag$: Reproduced by TranSTR, $\ddag$: Reproduced by MotionEpic.}
\label{tab:causal-vidqa}
\vspace{-3mm}
\end{table*}

Inspired by VFC~\cite{VFC}, we apply multi-stage visual hallucination mitigation using LLMs and MLLMs. First, we \textit{decompose} detailed region-level captions into multiple closed-ended questions that ask about the contents in the captions using LLMs. Then, we input these questions into MLLMs along with videos to validate whether the content is correct. In the third stage, we gather the questions not verified in the previous step and ask LLMs to remove the unverified contents in the original captions and re-generate them.

\subsection{GPT-Assited Region-level Instruction Data}
In the final step, building on the concept of caption-guided instruct-tuning data construction~\cite{LLaVA, Groma, VideoChatGPT, MiniGPT4}, we further process the captions to generate region-level video instructions. We utilize text-only GPT4~\cite{GPT4} to create region-specific question-answer pairs from the detailed captions, addressing various aspects of the captions. The samples include detailed descriptions, summaries, and general QAs for the specific regions. We provide a few in-context examples to enhance the quality of sample generation. The generated instructions cover both contextual (\eg, color, spatial positions) and temporal aspects (\eg, motions, actions).

\section{Experiments}
\label{sec:experiments}

\subsection{Implementation Details}
We leverage Llama-2~\cite{Llama2} as the language model, CLIP-ViT-L~\cite{CLIP} as the vision encoder, and a two-layer MLP for the projection layer.
For \featuremarkabbr{} generation, we set the number of \featuremarkabbr{} ($N_F$) to 100 with $C=256$ and employ a single linear layer as the projection layer.
The input image is resized to $336 \times 336$.
To embed the spatial \featuremarkabbr{} at the same resolution as the image tokens, we apply adaptive average pooling with a window size of 24.
Additionally, a single linear layer is applied for the auxiliary classifier, with $\mathcal{L}_{\text{aux}}$ coefficient $\alpha$ set to $0.05$.
For video data, we uniformly sample four frames.

\noindent \textbf{Image Pre-training.}
We adopt the pretraining recipe of RegionGPT~\cite{RGPT}, which utilizes a global image-text paired dataset in conjunction with an image-region dataset.

\noindent \textbf{Joint Fine-tuning.}
During fine-tuning, we employ a joint image-video training strategy. 
We include image-text paired datasets~\cite{LLaVA, FLAN} and image region-level dataset~\cite{RGPT}.
Additionally, we incorporate Visual Commonsense Reasoning (VCR)~\cite{VCR}, which combines recognition and cognitive-level understanding of complex scenes through a multiple-choice question-answering (QA) format.

For the region-level video dataset, we reformulate annotations from publicly available datasets, including Vid-STG~\cite{VidSTG}, Causal-VidQA~\cite{Causal-VidQA}, and ReVOS~\cite{VISA}.
We use captions paired with regions from Vid-STG and ReVOS to create region-level video captioning and Referring Expression Comprehension (REC)-style prompts.
Our \datasetabbr{} is also included to enhance the model's capacity to generate rich, detailed object captions.
In this stage, we train the model for one epoch, freezing only the vision encoder while allowing all other parameters to be updated.

\subsection{Quantitative Evaluation}
We report the performance of our method on both image and video domains.
All results are evaluated using the 7B language model with joint image-video trained weights unless stated otherwise.

\noindent \textbf{Region-level Video QA.}
In~\cref{tab:causal-vidqa}, we present our performance on the Causal-VidQA~\cite{Causal-VidQA} benchmark, designed to evaluate video question answering with a focus on causal reasoning across temporal and spatial dimensions.
Causal-VidQA comprises video clips paired with questions that assess the model’s ability to interpret events, interactions, and object relationships, requiring not only frame-level recognition but also temporal reasoning across sequences.

Our approach achieves state-of-the-art performance across all sub-tasks with a substantial margin, demonstrating robust region-level video reasoning.
In particular, our method achieves a significant performance margin in the Prediction (P) task, emphasizing advanced temporal reasoning capabilities. 
Additionally, we observe substantial improvements in the ``AR'' accuracy, which requires accurate reasoning for both answer and reason selections, underscoring the capability to justify its choices coherently.
Compared to MotionEpic~\cite{VoT}, which relies on frame-level proposals from every frame and an iterative scene graph encoding process, our method achieves strong performance \textit{without} requiring complete tracklet inputs or complex relational encoding modules.
Notably, we do not use the region's category as input, further emphasizing the effectiveness of \textit{direct region-language connection}.

\begin{table}[t]
\centering
\resizebox{\columnwidth}{!}{
\begin{tabular}{l|cccc}
\toprule
Models & Vid-STG & Extended-Elysium & Elysium& BensMOT  \\
\midrule
ViP-LLaVA~\cite{ViP-LLaVA} & 9.9 & 3.1 & 10.8 & 7.0 \\
LLaVA-OV~\cite{LLaVAOneVision} & 4.9  & 12.1  & 9.6  & 8.9  \\
Groma~\cite{Groma} & \underline{10.5}  & 2.1  & \underline{13.1}  & 6.0  \\
RegionGPT~\cite{RGPT} & 10.4 & \underline{14.4} & 13.0 & \underline{14.0}  \\
Elysium~\cite{Elysium} & 7.2 & 0.7 &  \textbf{19.1} & 1.1  \\
\midrule
\methodabbr{}  & \textbf{14.2}   & \textbf{19.3}  & 9.3  & \textbf{14.6}  \\
\bottomrule
\end{tabular}
}
\vspace{-2mm}
\caption{\textbf{Performance on the region-level video captioning task.} METEOR metric is reported.}
\label{tab:vid-reg-cap}
\vspace{-2mm}
\end{table}

\noindent \textbf{Region-level Video Captioning.}
We evaluate the region-level video captioning capability in \cref{tab:vid-reg-cap}.
For comparison, we include the zero-shot performance of image-based MLLMs~\cite{ViP-LLaVA, LLaVAOneVision, Groma, RGPT}, using only the first frame as input.
Since public benchmarks (Vid-STG~\cite{VidSTG} and Elysium~\cite{Elysium}) offer brief descriptions with limited temporal context (see \cref{tab:dataset_comparison}), we additionally generate detailed region-level captions on Elysium (Extended-Elysium) using GPT-4o~\cite{GPT4o}. Moreover, to evaluate the zero-shot captioning performance, we employ the BenSMOT dataset~\cite{BenSMOT}, which includes relatively detailed region-level video captions (40 words).
Our method demonstrates strong performance across datasets, except Elysium.
We attribute this to the nature of Elysium’s captions, which are typically brief nouns without motion-related information, favoring fine-tuned~\cite{Elysium} or image-specialized methods~\cite{ViP-LLaVA, LLaVAOneVision, Groma, RGPT}.

\begin{table}[t]
\centering
\small
\begin{tabular}{l|ccc}
\toprule
Model & Q $\rightarrow$ A & QA $\rightarrow$ R & Q $\rightarrow$ AR \\
\midrule
ViLBERT~\cite{ViLBERT}         & 72.4  & 74.5  & 54.0\\
Unicoder-VL~\cite{UniCoderVL}     & 72.6  & 74.5  & 54.5\\
VLBERT-L~\cite{VLBERT}        & 75.5  & 77.9  & 58.9\\
ERNIE-ViL-L~\cite{ERNIEViL}     & 78.5  & 83.4  & 65.8\\
VILLA-L~\cite{VILLA}         & 78.5  & 82.6  & 65.2\\
GPT4RoI~\cite{GPT4RoI}      & 87.4  & 89.6  & 78.6\\
ViP-LLaVA~\cite{ViP-LLaVA}   & \underline{87.7}  & \underline{89.8}  & \underline{78.9}\\ 
\midrule
\methodabbr{}          & \textbf{88.5} & \textbf{90.1} & \textbf{79.9}\\
\bottomrule
\end{tabular}
\vspace{-2mm}
\caption{\textbf{Accuracy (\%) on VCR validation set.} VCR includes three accuracy measures: (Q $\rightarrow$ A) selects the correct answer, (QA $\rightarrow$ R) selects the correct rationale given the correct answer, and (Q $\rightarrow$ AR) chooses both the correct answer and rationale. }
\label{tab:vcr}
\vspace{-2mm}
\end{table}

\noindent \textbf{Region-level Image QA.}
We report our model’s performance on Visual Commonsense Reasoning (VCR)~\cite{VCR} validation set (\cref{tab:vcr}).
VCR is designed to assess a model’s capability in commonsense reasoning within visual contexts, with each sample consisting of questions (Q) and multiple-choice options for both the answer (A) and rationale (R). 
Each question requires the model to select the correct answer and provide a rationale for supporting that answer. Following the protocols of GPT4RoI~\cite{GPT4RoI} and ViP-LLaVA~\cite{ViP-LLaVA}, the model is finetuned on VCR.

Our approach achieves state-of-the-art performance across all metrics, outperforming methods specifically designed for image-based tasks.
This highlights our method’s effectiveness in both image and video contexts using a unified architecture.
Furthermore, the proposed \featuremarkabbr{} facilitates a robust region-level connection between text prompts and visual context, enabling reasoning over complex, multi-region relationships.

\begin{table}[t]
\centering
\resizebox{\columnwidth}{!}{
\begin{tabular}{l|cccc}
\toprule
\multirow{2}{*}{\makecell{Models}} & \multicolumn{2}{c}{\makecell{RefCOCOg}} & \multicolumn{2}{c}{\makecell{Visual Genome}}  \\
& METEOR & CIDEr & METEOR & CIDEr \\ \midrule
GRiT~\cite{GRiT}$^\ddag$ & 15.2 & 71.6 & \textbf{17.1} & \underline{142.0} \\
ControlMLLM~\cite{ControlMLLM} & 14.0 & 59.8 & - & - \\
SLR~\cite{SLR} & 15.9 & 66.2 & - & - \\
Kosmos-2~\cite{kosmos2} & 14.1 & 62.3 & - & - \\
GLaMM~\cite{GLaMM} & 15.7 & 104.0 & \underline{17.0} & 127.0 \\
OMG-LLaVA~\cite{OMG-LLaVA} & 15.3 & - & - & - \\
ViP-LLaVA~\cite{ViP-LLaVA} & 16.6 & 105.9 & - & - \\
Groma~\cite{Groma} & 16.8 & 107.3 & 16.8$^\dagger$ & 137.4$^\dagger$\\
RegionGPT~\cite{RGPT} & \underline{16.9} & \textbf{109.9} & \underline{17.0} & \textbf{145.6} \\
\midrule
\methodabbr{}  & \textbf{17.0}  & \underline{109.7} & \underline{17.0} & 139.3 \\
\bottomrule
\end{tabular}
}
\vspace{-2mm}
\caption{\textbf{Performance on region-level image captioning on RefCOCOg and Visual Genome}, $\dagger$ denotes re-experimented results on the full evaluation set, $\ddag$ denotes the region caption specialists.}
\label{tab:reg-cap}
\vspace{-1mm}
\end{table}

\noindent \textbf{Region-level Image Captioning.}
We evaluate our region-level image captioning performance on RefCOCOg~\cite{RefCOCO} and Visual Genome~\cite{VisualGenome} in \cref{tab:reg-cap}.
Following prior works, we report METEOR~\cite{METEOR} and CIDEr~\cite{CIDEr} scores to measure captioning quality.
Note that Groma~\cite{Groma} reported results on a subset of the Visual Genome validation set; accordingly, we re-evaluated results on the complete set to ensure consistency with prior studies.
Across both datasets, our approach achieves comparable scores to those of methods specifically designed for the image domain, demonstrating its effectiveness at region-level captioning.

\begin{table}[t]
\centering
\resizebox{\linewidth}{!}{
\begin{tabular}{c|ccc|c}
\toprule
 & GPT4o~\cite{GPT4o}   & RegionGPT~\cite{RGPT} & Groma~\cite{Groma}  & \methodabbr{} \\ \midrule
val  & 74.08 & 86.44 & \underline{88.19}  & \textbf{95.99} \\
test & 72.23 & 86.96 & \underline{88.38}  & \textbf{95.88} \\
\bottomrule
\end{tabular}
}
\vspace{-2mm}
\caption{\textbf{Accuracy (\%) of REC on RefCOCOg val and test set}.}
\label{tab:image-rec-refcocog}
\vspace{-4mm}
\end{table}

\noindent \textbf{Image REC.}
To assess the region-level natural language referring capability, we adopt the REC evaluation protocol of RegionGPT~\cite{RGPT}.
For a concrete comparison, we evaluate GPT4o~\cite{GPT4o}, Groma~\cite{Groma}, and ours using the same region proposals~\cite{RGPT}.
As shown in \cref{tab:image-rec-refcocog}, our approach achieves strong performance with a significant margin.
The results demonstrate that our method establishes a robust and precise region-language connection by projecting token marks directly into visual features to indicate regions.

\subsection{Visualization Analysis}
\begin{figure}[t]
\begin{center}
\includegraphics[width=\linewidth]{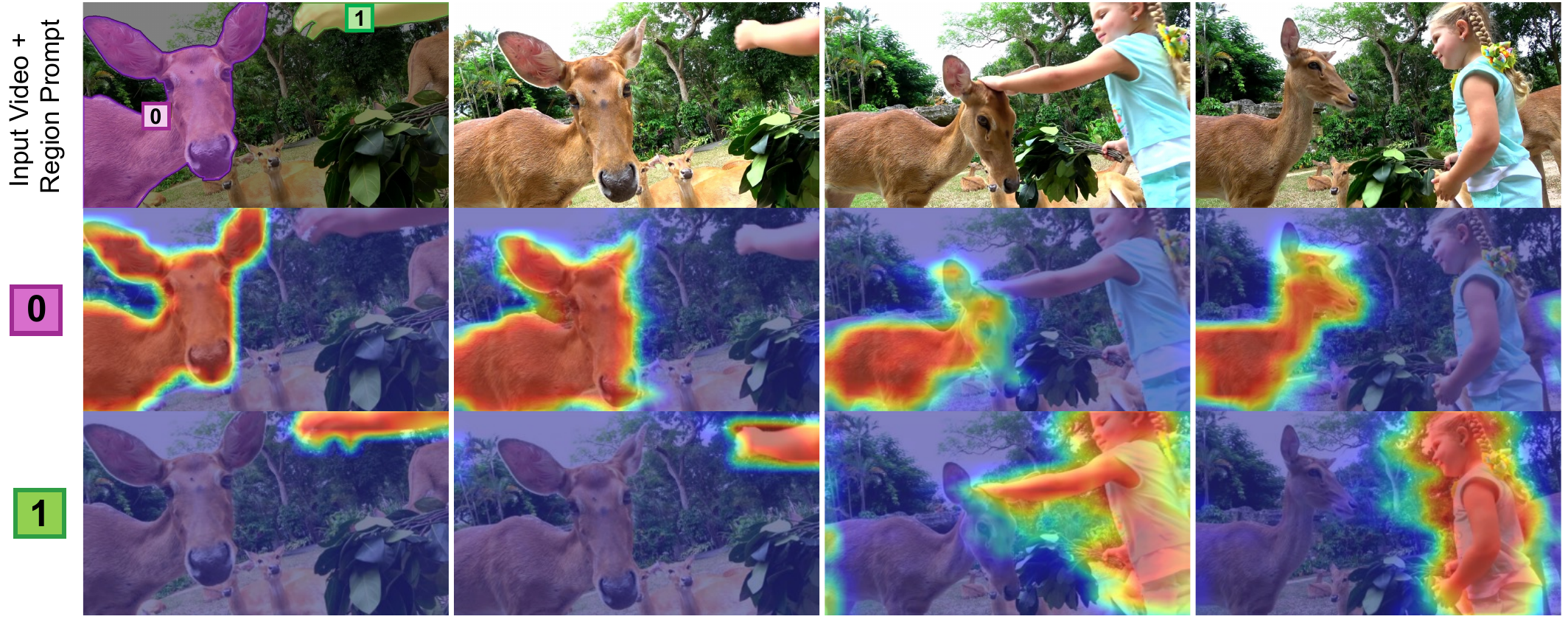}
\end{center}
\vspace{-5mm}
\caption{
\textbf{Heatmap of Temporal Region Guide Head outputs.}
}
\label{fig:aux_map}
\end{figure}

In \cref{fig:aux_map}, we analyze our Temporal Region Guide Head on the example shown in~\cref{fig:teaser_main} sourced from~\cite{VIPSeg}.
The first row displays input video frames, with region prompts and corresponding numbers overlaid on the initial frame for reference\footnote{Note that these masks and annotations are added for illustration only, and original image frames are used as input for the vision encoder.}.
We then assign randomly sampled \featuremarkabbr{} to each region prompt (deer and hand) in the first frame.
In the second and third rows, we visualize the output probabilities of the assigned \featuremarkabbr{} from the Region Guide Head as a heatmap.
In the first frame, where the regions are inputted as masks, the outputs accurately predict the region prompts.
In subsequent frames, even without the region prompt, the model successfully predicts the corresponding region by leveraging visual correlations across frames.

\subsection{Ablation Study}
\begin{table}[t]
\centering
\resizebox{\columnwidth}{!}{
\begin{tabular}{c|cc}
\toprule
Temporal Region Guide Head & Extended-Elysium & BenSMOT\\
\midrule
\checkmark & \textbf{19.3} & \textbf{14.6}\\
           & 16.2  & 12.2    \\
\bottomrule
\end{tabular}
}
\vspace{-2mm}
\caption{\textbf{Ablation study of temporal region guide head on video region-level captioning task.}}
\label{tab:ablation-aux-head}
\vspace{-3mm}
\end{table}
\noindent \textbf{Temporal Region Guide Head.}
We report the performance of our model trained without the auxiliary task in \cref{tab:ablation-aux-head}.
The results demonstrate that incorporating the auxiliary task further enhances the detailed object captions in videos.

\begin{table}[t]
\scalebox{0.9}{
\centering
\small
\begin{tabular}{l|cccc|c}
\toprule
\multirow{2}{*}{\makecell{Input \\ Frames}} &\multirow{2}{*}{D (A)}  & \multirow{2}{*}{E (A)} & \multirow{2}{*}{P (AR)} & \multirow{2}{*}{C (AR)} & \multirow{2}{*}{All}\\
& & & & & \\ \midrule
4 & 84.0 & \textbf{84.6} & \textbf{76.9} & \textbf{64.3} & \textbf{77.5}\\ \midrule
3 & \textbf{84.1}$_{\textcolor{green}{+0.1}}$ & 84.5$_{\textcolor{red}{-0.1}}$ & 76.1$_{\textcolor{red}{-0.8}}$ & 63.9$_{\textcolor{red}{-0.4}}$ & 77.2$_{\textcolor{red}{-0.3}}$\\
2 & 83.3$_{\textcolor{red}{-0.7}}$ & 84.5$_{\textcolor{red}{-0.1}}$ & 75.3$_{\textcolor{red}{-1.6}}$ & 62.2$_{\textcolor{red}{-2.1}}$ & 76.3$_{\textcolor{red}{-1.2}}$\\
1 & 82.7$_{\textcolor{red}{-1.3}}$ & 82.8$_{\textcolor{red}{-1.8}}$ & 71.7 $_{\textcolor{red}{-5.2}}$& 59.9$_{\textcolor{red}{-4.4}}$ & 74.3$_{\textcolor{red}{-3.2}}$\\
\bottomrule
\end{tabular}
}
\vspace{-2mm}
\caption{\textbf{Number of input frames ablation study on Causal-VidQA dataset.} We also show the performance gap from the four frames input.}
\label{tab:ablation-numframes}
\end{table}
\noindent \textbf{Number of Input Frames.}
We study the impact of the number of input frames on Causal-VidQA (\cref{tab:ablation-numframes}). 
The average accuracy across sub-tasks gradually decreases as the number of frames is reduced.
When comparing four and single-frame inputs, scene description (D) and explanation (E) show a modest accuracy.
In contrast, prediction and counterfactual tasks (P and C) exhibit a substantial decline, highlighting the importance of temporal information.

\noindent \textbf{Vision Encoding Method.}
In \cref{tab:ablation-resolution}, we investigate the impact of vision encoding methods inspired by recent conventions.
Our region classification results on the COCO dataset indicate that reducing the resolution of image tokens significantly diminishes region-level understanding.
This result aligns with recent studies in image-based MLLMs, which benefit from larger token sizes~\cite{LLaVA-Next, Eagle}. 
\begin{table}[t]
\centering
\resizebox{\columnwidth}{!}{
\begin{tabular}{lcc|cc}
\toprule
Vision encoder & Downscale & Token size & mAP & Acc  \\
\midrule
CLIP-ViT-L-336 & \xmark & 24$\times$24 & \textbf{68.2} & \textbf{78.6} \\
SigLip-SO400M-384 & Pixel unshuffle  & 14$\times$14  & 59.7  & 70.0  \\
\bottomrule
\end{tabular}
}
\vspace{-2mm}
\caption{\textbf{Vision encoding methods ablation study on COCO classification}.}
\label{tab:ablation-resolution}
\vspace{-3mm}
\end{table}

\section{Limitation}
\label{sec:limitation}
As shown in our ablation studies, robust region-level video understanding requires dense spatial and temporal visual information.
However, similar to existing region-level video understanding models~\cite{Elysium, Merlin}, achieving high fidelity in both dimensions is challenging within our current framework. 
Using a four-frame video input may not fully capture the complexity of diverse, real-world scenarios. 
Extending region-level understanding to accommodate long-form videos is a promising direction for future research.

\section{Conclusion}
\label{sec:conclusion}
We present \methodabbr{}, a multimodal LLM with region-specific comprehension capabilities for both images and videos.
The main idea is to use \featuremarkabbr{}, a set of tokens that bridges the language and spatio-temporal visual tokens at region-level.
For robust region understanding in videos, we introduce an auxiliary task tailored for video inputs.
To further enhance detailed region-description capabilities in videos, we introduce \datasetabbr{}, a new dataset re-annotated with GPT-4o on publicly sourced videos.
With its intuitive design, \methodabbr{} surpasses existing methods on challenging visual commonsense reasoning benchmarks in both image-based (VCR) and video-based (Causal-VidQA) tasks.
Additionally, it demonstrates strong performance on diverse region-level understanding tasks, including captioning and referring expression comprehension.

\paragraph{Acknowledgements.}
We would like to thank Qiushan Guo for assistance in setting up the project. We also appreciate the NVIDIA VILA team for the efforts in developing a robust framework that greatly facilitated our research.

{
    \small
    \bibliographystyle{ieeenat_fullname}
    \bibliography{main}
}

\clearpage
\setcounter{page}{1}
\maketitlesupplementary

In this supplementary material, we provide additional insights and details to support the main paper.
In~\cref{sec:supp-implementation}, we outline the implementation details, including dataset setups and training strategies. 
\cref{sec:supp-quantitive} presents extended quantitative results, offering further validation of our method's performance.
To provide a more comprehensive understanding, \cref{sec:supp-qualitative} includes a set of qualitative results, showcasing visual examples that demonstrate our model's effectiveness in diverse scenarios.
\cref{sec:supp-vis-analysis} provides a visualization analysis, including an exploration of the limitations of our model.
In~\cref{sec:supp-dataset}, we offer additional details about our newly constructed dataset, \datasetabbr. Finally, in~\cref{sec:supp-ethics}, we discuss important ethical considerations.

\section{Implementation Details}
\label{sec:supp-implementation}
We further provide detailed setups for the image dataset.
Our training consists of two stages: image pre-training, followed by image-video joint fine-tuning, which incorporates our newly proposed region-level video instruction dataset, \datasetabbr.
For pretraining, we utilize a image-text paired dataset alongside an image-region pre-training dataset as proposed in RegionGPT~\cite{RGPT}.
Specifically, following VILA~\cite{VILA}, we use interleaved image-text data~\cite{MMC4} and conventional image-text pairs~\cite{COYO}.
The image-region pre-training dataset includes Visual Genome~\cite{VisualGenome}, RefCOCOg~\cite{RefCOCO}, and V3Det~\cite{V3Det}.
For the image region-level dataset in the joint fine-tuning stage, we utilize the RegionGPT's fine-tuning dataset, which includes the ReCapD, a high-quality, GPT-assisted region-aware image dataset.

During pre-training, all learnable parameters, except those in the visual encoder and the language model, are trained for one epoch.
In the fine-tuning stage, the model is trained for one epoch, with the vision encoder frozen while allowing all other parameters to be updated.
We use a batch size of 16 and a learning rate of $5 \times 10^{-5}$.
The entire training process, including both pre-training and fine-tuning stages, is completed within 24 hours using 8 nodes of 8$\times$A100 GPUs.

\section{More Quantitative Results}
\label{sec:supp-quantitive}
\subsection{VCR}
\begin{table}[t]
\centering
\small
\resizebox{\columnwidth}{!}{
\begin{tabular}{ccc|ccc}
\toprule
 Fine-tuning & Category & Region & Q $\rightarrow$ A & QA $\rightarrow$ R & Q $\rightarrow$ AR \\
\midrule
 \checkmark & \checkmark & mask & \textbf{88.5} & \textbf{90.1} & \textbf{79.9}\\\midrule
\checkmark &            & mask & 88.2 & \textbf{90.1} & 79.8 \\
\checkmark &            & bbox & 87.7 & 90.0 & 79.2\\
           &            & bbox & 85.8 & 87.5 & 75.3\\
\bottomrule
\end{tabular}
}
\caption{\textbf{Accuracy (\%) on VCR with different settings.} }
\label{tab:supp-vcr}
\vspace{-3mm}
\end{table}
In~\cref{tab:supp-vcr}, we present evaluation results on VCR~\cite{VCR} validation set under various settings, including category name usage, different region prompt representations (\eg, bounding boxes and masks), and dataset-specific fine-tuning.
The first row in the table corresponds to the results presented in the main paper, following the same evaluation protocol as existing methods~\cite{GPT4RoI, ViP-LLaVA}.

\noindent \textbf{Category Name Usage.}
Manually specifying category names for each target can be inefficient in real-world, user-interactive applications. 
For instance, consider the question ``Why is \texttt{<region1>} moving towards \texttt{<region2>}?'' paired with region prompts that include category names, such as \texttt{<region1>:person} and \texttt{<region2>:animal}.
Incorporating these categories transforms the question into, ``Why is person \texttt{<region1>} moving towards animal \texttt{<region2>}?''

In the second row, we evaluate our model’s performance without utilizing such category names for the provided region prompts.
While the results show a slight decrease in performance, they highlight that \methodabbr{} internalizes the semantic knowledge of regions within the question. 
This is achieved by leveraging visual context through a \textit{direct region-language connection} using \featuremarkabbr{}.

Notably, unlike RoI-based methods~\cite{RGPT, GPT4RoI}, which inject explicit visual features into the text prompt to embed semantic knowledge, \methodabbr{} accomplishes this by utilizing randomly sampled \featuremarkabbr{} within the text prompt. 
This highlights its ability to generalize and infer semantics directly from the visual context itself.

\noindent \textbf{Region Prompt Representation.}
VCR provides two types of localization information for targets: bounding box coordinates and polygon-style segmentations.
In the third row, we report the result using bounding box coordinates as the region prompt representation, which offers a less fine-grained target representation compared to masks.
The overall performance shows a marginal decrease compared to the results obtained using masks, highlighting the flexibility of our framework in accommodating dual region prompt types.

\noindent \textbf{Task-specific Fine-tuning.}
In the last row, we evaluate the performance of \methodabbr{} without VCR-specific fine-tuning.
We use joint image-video trained weights, maintaining consistency with the settings used for other quantitative results presented in the main paper.
The results here demonstrate that \methodabbr{} retains competitive performance even without task-specific fine-tuning, showcasing its robustness and broader applicability.

\subsection{Effectiveness of \datasetabbr{}}
\begin{table}[t]
\centering
\resizebox{\columnwidth}{!}{
\begin{tabular}{c|p{1.1cm}p{1.1cm}p{1.1cm}p{1.1cm}|p{1.1cm}}
\toprule
\multirow{2}{*}{\makecell{RegVID \\ -300k}} &\multicolumn{4}{c|}{Causal-VidQA} & \multirow{2}{*}{\makecell{Extended \\ -Elysium}}\\
& D (A) & E (A) & P (AR) & C (AR) & \\ \midrule
\checkmark & \textbf{84.0} & \textbf{84.6} & \textbf{76.9} & \textbf{64.3} & \textbf{19.3}\\ 
           & 82.7$_{\textcolor{red}{-1.3}}$ & 84.4$_{\textcolor{red}{-0.2}}$ & \textbf{76.9} & 63.5$_{\textcolor{red}{-0.8}}$ & 13.5$_{\textcolor{red}{-5.8}}$\\

\bottomrule
\end{tabular}
}
\caption{\textbf{Effectiveness of RegVID-300k.} Accuracy for Causal-VidQA and METOR scores for Extended-Elysium are reported.}
\label{tab:supp-reg-vid}
\vspace{-2mm}
\end{table}
In~\cref{tab:supp-reg-vid}, we report the effectiveness of \datasetabbr{}, our new region-level video instruction dataset, on Causal-VidQA~\cite{Causal-VidQA} and Extended-Elysium.
The first row is the results reported in the main paper, obtained using the fine-tuned weight that incorporates the new dataset.
In contrast, the second row shows results where the model is trained under the same settings but without including the new dataset, relying solely on annotations from public video datasets.
The results underscore the effectiveness of \datasetabbr{}, as evidenced by the performance degradation, including the Description (D) task within Causal-VidQA ($-1.3$) and Extended-Elysium ($-5.8$).
In particular, the large performance drop in Extended-Elysium highlights the contribution of our new dataset’s diverse and detailed descriptions, which enhance the model’s region-level captioning capability in the video domain.

\subsection{Hallucination}
\begin{table*}[t]
\begin{center}
\setlength{\tabcolsep}{3.5pt}
\resizebox{0.95\textwidth}{!}{
\begin{tabular}{l|l|cccccccc}
\toprule
POPE & Metrics & \methodabbr{} & RegionGPT~\cite{RGPT} & Shikra \cite{Shikra} & InstructBLIP \cite{InstructBLIP} & MiniGPT4 \cite{MiniGPT4} & LLaVA \cite{LLaVA} & MM-GPT \cite{mmgpt} & mPLUG-Owl \cite{mplugowl}  \\
\midrule
\multirow{5}{*}{Random}&Accuracy ($\uparrow$) & 88.76 & 87.80 & 86.90 & 88.73 & 77.83 & 86.00 & 50.03 & 53.30 \\
& Precision ($\uparrow$) & 91.15 & 97.75 & 94.40 & 85.08 & 75.38 & 87.50 & 50.02 & 51.71 \\
& Recall ($\uparrow$) & 86.60 & 78.13 & 79.26 &  93.33 & 82.67 & 84.00 & 100.00 & 99.53 \\
& F1 Score ($\uparrow$) & 88.82 & 86.85  & 86.19 & 89.29 & 78.86 & 85.71 & 66.68 & 68.06 \\
& Yes & 48.96 & 41.20 & 43.26 & 55.20 & 54.83 & 48.00 & 99.97 & 96.23 \\
\midrule
\multirow{5}{*}{Popular}& Accuracy ($\uparrow$) & 86.63 & 87.20 & 83.97 & 81.37 & 68.30 & 76.67 & 50.00 & 50.63 \\
& Precision ($\uparrow$) & 86.20 & 95.44 & 87.55 & 75.07 & 64.27 & 72.22 & 50.00 & 50.32 \\
& Recall ($\uparrow$) & 86.60 & 78.13  & 79.20 & 93.33 & 82.40 & 86.67 & 100.00 & 99.27 \\
& F1 Score ($\uparrow$) & 86.40 & 85.92  & 83.16 & 84.35 & 72.21 & 78.79 & 66.67 & 66.79 \\
& Yes & 50.23 & 40.93 & 45.23 & 62.57 & 64.10 & 60.00 & 100.00 & 98.63 \\
\midrule
\multirow{5}{*}{Adversarial}& Accuracy ($\uparrow$) & 81.67 & 85.67 &  83.10 & 74.37 & 66.60 & 73.33 & 50.00 & 50.67 \\
& Precision ($\uparrow$) & 78.82 & 91.99 & 85.60 & 67.67 & 62.45 & 69.02 & 50.00 & 50.34 \\
& Recall ($\uparrow$) & 86.60 & 78.13 &  79.60 & 93.33 & 83.27 & 84.67 & 100.00 & 99.33 \\
& F1 Score ($\uparrow$) & 82.53 & 84.50  & 82.49 & 78.45 & 71.37 & 66.32 & 66.67 & 66.82 \\
& Yes & 54.93 & 42.47 & 46.50 & 68.97 & 66.67 & 61.33 & 100.00 & 98.67\\
\bottomrule
\end{tabular}
}

\caption{ \textbf{Results on the object hallucination benchmark using the POPE evaluation pipeline on MSCOCO.}}
\label{tab:supp-pope}
\end{center}
\vspace{-3mm}
\end{table*}
Following the POPE~\cite{POPE} evaluation protocol, we report object hallucination evaluation results in~\cref{tab:supp-pope}.
Performance is evaluated on the MSCOCO dataset under three settings: Random, Popular, and Adversarial.
In the Random setting, our method achieves the best Accuracy and a competitive F1 Score, trailing the best-performing model (InstructBLIP~\cite{InstructBLIP}) by only $-0.47$.
In the Popular setting, our Accuracy is marginally below the best-performing model (RegionGPT) by $-0.57$ while achieving the highest F1 Score among all methods.
Finally, in the Adversarial setting, both Accuracy and F1 scores are lower than RegionGPT, with differences of $-4.0$ and $-1.97$, respectively.
Overall, our method demonstrates competitive performance compared to the image-specialized counterparts.

\section{More Qualitative Results}
\label{sec:supp-qualitative}
We showcase visualizations of \methodabbr{}’s region-level understanding capabilities across both image and video scenarios.
For all examples, we use a jointly fine-tuned model using the 7B language model.
The inputs, including mask proposals, are sourced from the VIPSeg~\cite{VIPSeg}, which is not included in the training dataset (see~\cref{tab:data_source}).
\subsection{Video Region-level Understanding}
\noindent \textbf{Brief Region-level Captioning.}
We present \methodabbr{}’s brief region-level video captioning capability in~\cref{tab:supp_video_short_caption}.
The results show that \methodabbr{} effectively generates concise captions for the given prompts, incorporating visual context across the entire sequence.
For example, in the second case, the caption for the red backpack (\texttt{<region0>}) notes that it is unzipped, a detail observable from the third frame.

\noindent \textbf{Detailed Region-level Captioning.}
\cref{tab:supp_video_long_caption} visualizes examples of detailed region-level video captioning.
Key information is highlighted in light orange color.
The outputs effectively capture transitions in the target’s location and motion across the sequence.
For instance, in the first case, \methodabbr{} describes the pillow (\texttt{<region0>}) as initially lying flat, later being lifted and adjusted by the person.

\noindent \textbf{Region-level QA.}
In~\cref{tab:supp_video_qa}, we present various examples of region-level question-answering in videos.
\methodabbr{} generates answers to diverse questions, including those about multi-object relationships and the location of a target within a sequence (\eg, see the second example).

\subsection{Image Region-level Understanding}
\noindent \textbf{Region-level Captioning.}
In~\cref{tab:supp-img-captioning}, we demonstrate \methodabbr{}'s captioning capabilities, including both brief and detailed descriptions.
The results highlight the model’s ability to describe specific details about the target object (\eg, the names of instruments), including its visual features such as color, material, and size.

\noindent \textbf{Region-level QA.}
In~\cref{tab:supp-img-qa}, we present a question-and-answering example that reflects a visual commonsense reasoning scenario.
In this example, \methodabbr{} successfully generates answers by considering the commonsense context within the scene.
Specifically, the output recognizes that wearing a helmet in an office is unusual, explains why, and suggests multiple possible reasons for such a case.

\noindent \textbf{Referring Expression Comprehension (REC).}
In~\cref{tab:supp-img-rec-comp} and~\cref{tab:supp-img-rec-comp-2}, we provide a visualization comparison with existing image-specialized methods~\cite{RGPT, Groma} on REC task using the RefCOCOg~\cite{RefCOCO}.
Among multiple proposals in the image, \methodabbr{} accurately identifies the most suitable one for the given caption.
As shown in the examples, the challenge in this task lies in the presence of multiple objects with similar appearances in the scene, requiring a precise understanding of the spatial location of each visual element.

Our understanding of \methodabbr{}’s robust REC capacity is that \featuremarkabbr{} establishes a strong ability to refer to visual tokens at the pixel level.
This enables better differentiation between visually similar objects compared to RoI-based methods, which primarily rely on similarity using averaged appearances to locate the target region prompt.

\section{Visualization Analysis}
\label{sec:supp-vis-analysis}
\noindent \textbf{Elysium dataset.}
As discussed in the main paper, we observe that the video region-level captioning dataset from Elysium~\cite{Elysium} is limited in capturing the rich information within video sequences.
In~\cref{tab:supp_elysium}, we illustrate Elysium’s original inputs with ground-truth captions alongside responses generated by our model.
From the examples, we note that: 1) the videos typically exhibit monotonous motion with limited dynamics, 2) the dataset primarily contains single dominant objects, which restricts its ability to evaluate multi-object scenarios, 3) the original ground truth captions are closer to noun-based descriptions rather than full captions.
To facilitate better evaluation, we refined its annotations (referred to as Extended-Elysium in our main paper) using the same pipeline as our dataset curation.

\noindent \textbf{Failure case.}
In~\cref{tab:supp_video_failure}, the first example shows that the current design occasionally exhibits unstable performance to understand small objects.
Additionally, since we employ frame-independent visual encoding without explicit temporal embeddings, our model has limited capability in interpreting the direction of objects.

\noindent \textbf{Occluded objects.}
In~\cref{tab:supp_occluded_case}, we analyze a heatmap of the Temporal Region Guide Head in diverse scenarios of occluded objects from~\cite{OViS}.

\section{\datasetabbr{}}
\label{sec:supp-dataset}
\begin{table*}[tb]
\centering
\resizebox{1.8\columnwidth}{!}{
\begin{tabular}{l|ccc}
\toprule
Dataset & Original Task & Region Annotation & Number of Used Videos \\ \midrule
MeViS~\cite{MeViS} & \multirow{2}{*}{Referring video object segmentation} & \multirow{2}{*}{masklet} & 1.6k \\
Ref-YTVOS~\cite{RefYTVOS} &  & & 3.4k \\ \midrule
Vid-STG~\cite{VidSTG} & \multirow{2}{*}{Spatio-Temporal Grounding} & \multirow{2}{*}{tracklet} & 5.3k \\ 
HC-STVG~\cite{HCSTVG} & & & 10.1k \\\midrule
BDD100k~\cite{BDD100k} & \multirow{2}{*}{Object Tracking} &  \multirow{2}{*}{tracklet} & 4.1k \\
ElysiumTrack-1M~\cite{Elysium} &  & & 10k \\\midrule
LV-VIS~\cite{LVVIS} & Open-vocabulary video instance segmentation & masklet & 3k \\\midrule
ActivityNet-Entities~\cite{ANetEntities} & Dense Video Grounding & bounding box & 38.8k \\\midrule
AVA-Kinetics~\cite{AVAKinetics} & Action Localization & bounding box & 13.4k \\\midrule
Action Genome~\cite{ActionGenome} & Action Graph & bounding box & 8.4k \\\bottomrule
\end{tabular}}
\caption{\textbf{Source video dataset summary table.} We collect the videos and corresponding region annotations from diverse datasets. }
\label{tab:data_source}
\end{table*}
\subsection{Data Preparation}
\begin{figure}[tb]
\centering
\resizebox{\columnwidth}{!}{
\begin{tabular}{cc}
\begin{minipage}{0.5\columnwidth}
    \begin{center}
        \includegraphics[clip, width=\columnwidth]{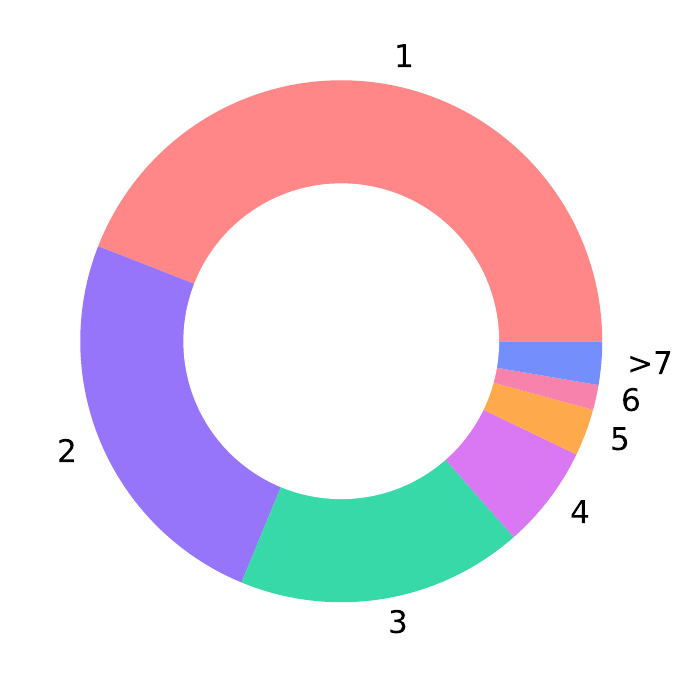} 
        (a) Number of regions
    \end{center}
\end{minipage}
&
\begin{minipage}{0.5\columnwidth}
    \begin{center}
        \includegraphics[clip, width=\columnwidth]{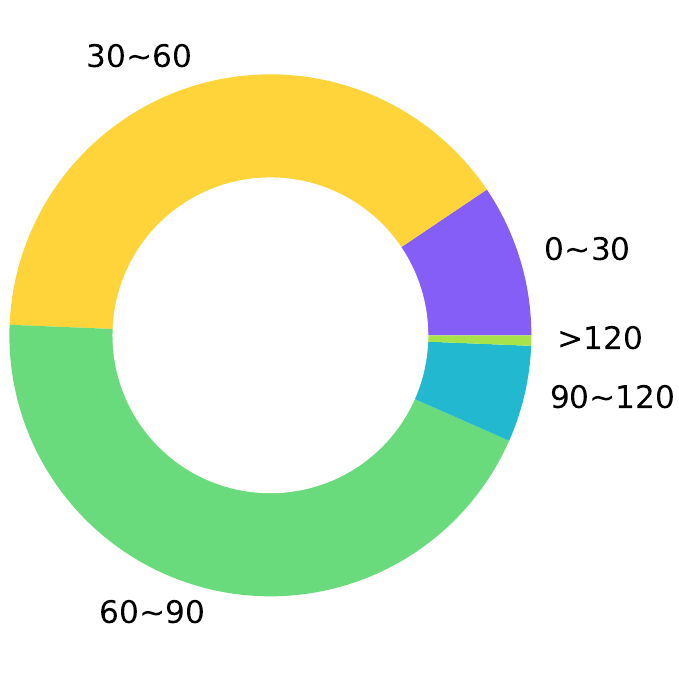} 
        (b) Caption length
    \end{center}
\end{minipage}
\end{tabular}}
\caption{\textbf{\datasetabbr{} statistics.} (a) The dataset contains multiple regions in videos. (b) The captions primarily range from 30 to 60 words, resulting in diverse instruction-following samples.}
\label{fig:data_stats}
\vspace{-3mm}
\end{figure}
We select public datasets containing videos and manually annotated regions (\eg, masklets, tracklets, single-frame bounding boxes) along with their associated region nouns. The full list of data sources, number of videos, and tasks used in the original datasets is summarized in \cref{tab:data_source}. Since each dataset inherently contains biases due to the dataset construction process—--such as video selection, annotation rules, or targeted scenes--—we diversify the source datasets based on the targeted tasks. We use only a subset of the ElysiumTrack-1M~\cite{Elysium} and AVA-Kinetics~\cite{AVAKinetics} datasets to diversify the video sources instead of increasing the videos within a single dataset. Because the original MeViS dataset~\cite{MeViS} does not provide objects' nouns, we generate them by using GPT-4~\cite{GPT4} to extract target nouns from annotated referring sentences for each object. Also, since the ActivityNet-Entities dataset~\cite{ANetEntities} consists of relatively long frames and includes scene changes, we divide each video into shorter clips based on the ground-truth event segments. Detailed statistics are visualized in \cref{fig:data_stats}. Our dataset includes multiple regions within each video, with detailed descriptions for each region.

\subsection{Data Visualization}
The QA instruction-following samples and their corresponding videos are presented in \cref{tab:data_sample_1,tab:data_sample_2}. These visualizations demonstrate the automatic generation of diverse and high-quality question-and-answer pairs related to the regions within the videos. The automated, multi-stage approach used to generate these region-level instruction-following samples is detailed in the following sections.

\subsection{GPT4o-Assisted Region-level Captioning}
Since not all datasets have annotated object masks per frame (\ie, masklets), we apply SAM~\cite{SAM} with ViT-H backbone~\cite{ViT} to the datasets that only contain the tracklets, such as Vid-STG~\cite{VidSTG}, HC-STVG~\cite{HCSTVG}, BDD100k~\cite{BDD100k}, and ElysiumTrack-1M~\cite{Elysium}. Moreover, for the datasets that only contain the single frame bounding box, such as ActivityNet-Entities~\cite{ANetEntities}, AVA-Kinetics~\cite{AVAKinetics}, and Action Genome~\cite{ActionGenome}, we apply SAM2~\cite{SAM2} to propagate the annotated bounding box to before and after the annotated frames to obtain the masklets. We followed the official code of SoM~\cite{SoM} to overlay the masklets onto the image frame-by-frame. Due to the context length and the budget limitation, we randomly subsample 16 frames from the video and resize to $334 \times 334$ to input to GPT4o with the text prompt~\cref{tab:gpt4o_prompt}. We input SoM-processed videos along with the object nouns to GPT4o to obtain the region-level captions. We employ \textbf{\texttt{gpt-4o-2024-08-06}} model in the paper.

\subsection{Visual Hallucination Mitigation}
We employ LLaMA-3.1-8B~\cite{Llama3} for the LLM in the first and third stages (decomposing the caption into questions and re-generating the caption) and Qwen2-VL-7B~\cite{Qwen2VL} for MLLMs for the second step (multi-modal fact-checking). 
In the first step, the multiple questions, which ask about the contents in the captions, are generated from the region-level caption using the text prompt~\cref{tab:decomp_prompt}. As mentioned in the text prompt, we query the LLM only to include the questions that ask the contents in the input captions. In the second step, we input the generated questions into the Qwen2-VL with the original videos to verify the contents. We ask Qwen2-VL to answer the multi-choice closed-ended questions (\cref{tab:fact_checking_prompt}). We input $8$ video frames due to the context length. We collect the questions whose answers are ``(D) Definitely No'' and input these questions with the original captions to remove the non-validated contents from the original captions and re-generate them (\cref{tab:regenerate_cap}).

\subsection{Region-level Instruction Data Generation}
From the refined region-level captions, we generate diverse instruction-following data. Three types of instruction-following data are generated: detailed descriptions, brief descriptions, and general conversations. 

\noindent \textbf{Region-level Detailed Description.} We directly use the refined captions for each region as an answer. For the instructions, we augmented the input instruction prompt using GPT4~\cite{GPT4} from the original text prompt (\texttt{Describe the region in detail}), resulting in diverse input text prompts to enhance the LLM's adaptability to users' various text prompts (\cref{tab:detailed_caption_templates}). For each region-level caption, we randomly select one of the instruction samples from the list to construct the question-and-answer pairs. 

\noindent \textbf{Region-level Brief Description.} We employ GPT4~\cite{GPT4} to summarize the region-level detailed caption using the text prompt~\cref{tab:summary_prompt} and generate the region-level short descriptions. Similar to the region-level detailed description, we also diversify the input instruction from (\texttt{Briefly describe the region}), allowing users' diverse input questions (\cref{tab:brief_caption_templates}), and the question is randomly selected from the list to construct the question-and-answer pairs for the region-level brief description task. 

\noindent \textbf{Region-level Conversation.} Inspired by LLaVA~\cite{LLaVA}, we generate diverse region-specific question-and-answer pairs from the region-level captions using GPT4. The prompt to GPT4 is shown in \cref{tab:conv_prompt}. Since the captions contain diverse aspects of the regions in the video, the generated questions also cover the diverse aspects of them.

\section{Ethics Concerns}
\label{sec:supp-ethics}
We utilize the pre-trained weights of the LLM (Llama-2~\cite{Llama2}) and fine-tune them using publicly available image and video datasets, along with our re-annotated datasets. Consequently, societal biases present in the original LLM model weights, which are derived from large-scale web text corpora, may be inherited in the output response.
\begin{table*}[t]
\centering
\resizebox{0.98\linewidth}{!}{
\begin{tcolorbox}[colback=white!100]
\centering
\begin{minipage}{\textwidth}
    \centering
    \vspace{2mm}
    \includegraphics[width=0.95\textwidth]{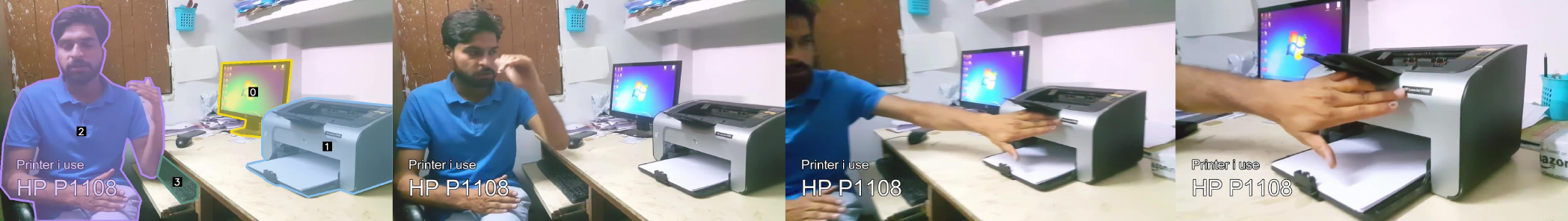} \\[0.5em]
    \vspace{1mm}
    \begin{tabular}{p{0.95\textwidth}}
    \textbf{User}: Give a brief interpretation of what stands out in \hlyellow{\texttt{<region0>}}.\\
    \textbf{Omni-RGPT}: The monitor is on a desk with a printer, indicating a workspace setup.\\
    \rule{15.6cm}{0.5pt} \\
    \textbf{User}: Provide a short summary of what’s visible in \hlblue{\texttt{<region1>}}.\\
    \textbf{Omni-RGPT}: The printer is a sleek, modern device on a desk in a home office setting.\\
    \rule{15.6cm}{0.5pt} \\
    \textbf{User}: Describe the notable objects or subjects you can see in \hlpurple{\texttt{<region2>}}.\\
    \textbf{Omni-RGPT}: A person in a blue shirt operates a printer in a home office setting.\\
    \rule{15.6cm}{0.5pt} \\
    \textbf{User}: How would you quickly explain what is featured in \hlgreen{\texttt{<region3>}}?\\
    \textbf{Omni-RGPT}: The keyboard is on a desk near a printer, used for computer tasks.\\
    \vspace{3mm}
    \includegraphics[width=0.95\textwidth]{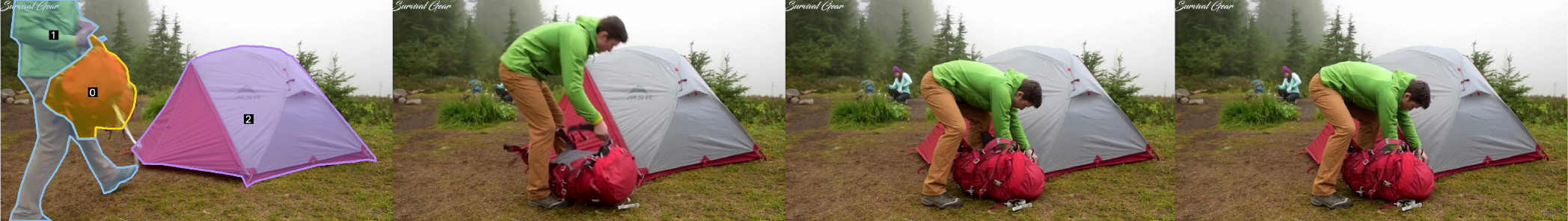} \\
    \vspace{1mm}
    \textbf{User}: Provide a concise rundown of what is visible in \hlyellow{\texttt{<region0>}}.\\
    \textbf{Omni-RGPT}: A red backpack is being unzipped and organized near a tent in a misty outdoor setting.\\
    \rule{15.6cm}{0.5pt} \\
    \textbf{User}: Describe the notable objects or subjects you can see in \hlblue{\texttt{<region1>}}.\\
    \textbf{Omni-RGPT}: A person in a green jacket and brown pants is organizing a red backpack near a tent in a foggy outdoor setting.\\
    \rule{15.6cm}{0.5pt} \\
    \textbf{User}: What are the key points of focus in this section \hlpurple{\texttt{<region2>}}?\\
    \textbf{Omni-RGPT}: A red and gray tent is set up on a grassy area in a misty forest.\\
    \vspace{3mm}
    \includegraphics[width=0.95\textwidth]{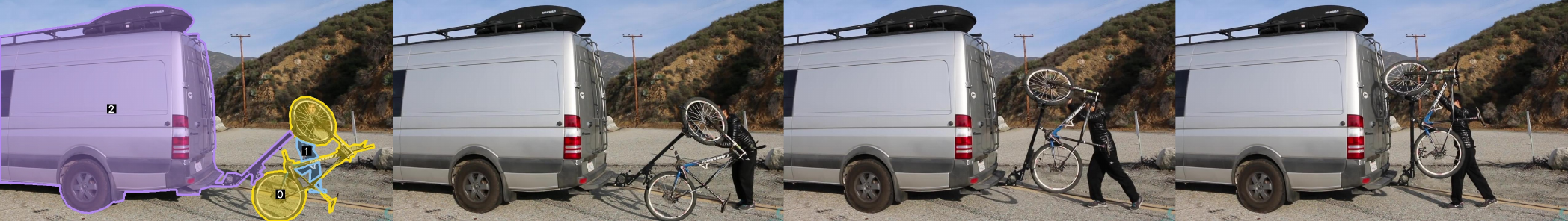} \\
    \vspace{1mm}
    \textbf{User}: Create a short label for the visible content in \hlyellow{\texttt{<region0>}}.\\
    \textbf{Omni-RGPT}: A bicycle is being lifted onto a rack on a van in a parking lot.\\
    \rule{15.6cm}{0.5pt} \\
    \textbf{User}: Draft a short title for the video content enclosed by \hlblue{\texttt{<region1>}}.\\
    \textbf{Omni-RGPT}: A person is securing a bicycle to a trailer attached to a van in a rural setting.\\
    \rule{15.6cm}{0.5pt} \\
    \textbf{User}: Provide a concise label for the highlighted \hlpurple{\texttt{<region2>}}.\\
    \textbf{Omni-RGPT}: A silver van is parked on a gravel surface with a bicycle attached to its rear.\\
    \end{tabular}
\end{minipage}
\end{tcolorbox}
}
\caption{\bfseries Qualitative visualization of brief region-level video captioning capability.}
\label{tab:supp_video_short_caption}
\end{table*}
\begin{table*}[t]
\centering
\resizebox{0.98\linewidth}{!}{
\begin{tcolorbox}[colback=white!100]
\centering
\begin{minipage}{\textwidth}
    \centering
    \vspace{4mm}
    \includegraphics[width=0.95\textwidth]{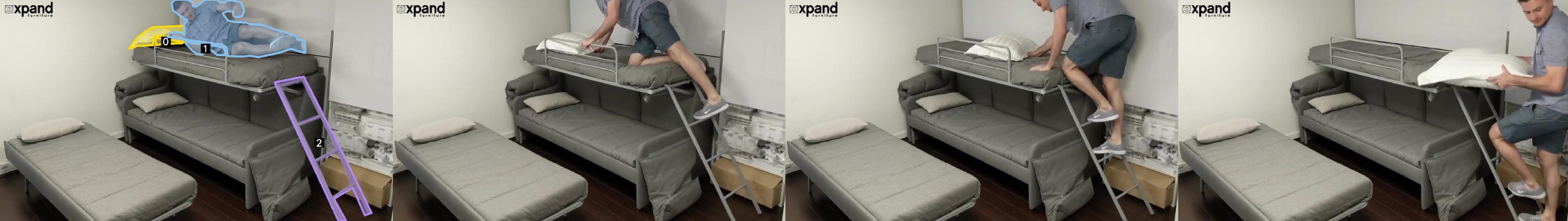} \\[0.5em]
    \vspace{1mm}
    \begin{tabular}{p{0.95\textwidth}}
    \textbf{User}: Give an extensive description of the area labeled \hlyellow{\texttt{<region0>}}, paying attention to textures, colors, movements, and notable details. \\[0.5em]
    \textbf{Omni-RGPT}: The pillow is positioned on the top bunk of a bunk bed. \hl{It is initially lying flat, but as the sequence progresses, it is lifted and adjusted by the person.} The pillow is white and appears soft, contrasting with the gray bedding. It is situated in a minimalist room with a wooden floor and white walls, providing a simple and clean backdrop.\\[0.5em]
    \rule{15.6cm}{0.5pt} \\
    \textbf{User}: Explain the visible details of the subject in \hlblue{\texttt{<region1>}}, including its appearance, movements, and any distinguishing traits. \\[0.5em]
    \textbf{Omni-RGPT}: The person is positioned on a bunk bed, initially lying on the top mattress. They are wearing a light-colored shirt and shorts, and appear to be adjusting their position. The room is minimalistic, with a plain wall and a few pieces of furniture, including a ladder leading to the top bunk. \hl{The person is seen moving from a reclined position to sitting up, then standing, indicating a transition from rest to activity.} The setting suggests a casual, relaxed environment. \\
    \rule{15.6cm}{0.5pt} \\
    \textbf{User}: Explain the key attributes and features of the subject found in \hlpurple{\texttt{<region2>}} with in-depth detail. \\
    \textbf{Omni-RGPT}: The ladder is positioned on the right side of the room, attached to the bunk bed. It is metallic and extends upwards, providing access to the upper bunk. \hl{The ladder is stationary, serving as a functional piece of furniture in the room.}\\
    \vspace{2mm}
    \includegraphics[width=0.95\textwidth]{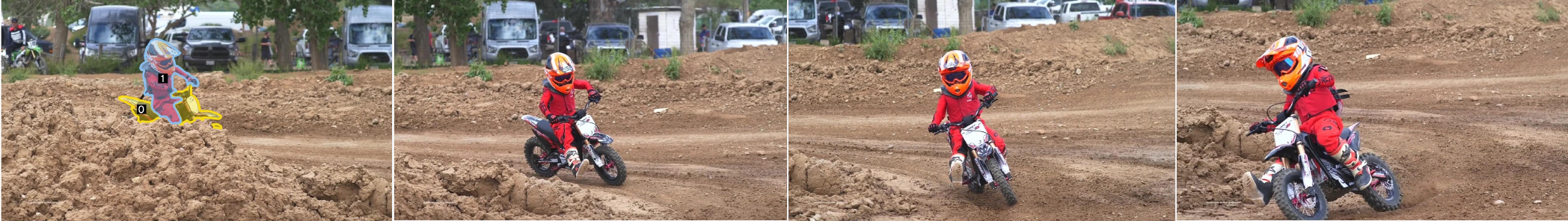} \\
    \vspace{1mm}
    \textbf{User}: Create a thorough, descriptive caption for the portion of the video located in \hlyellow{\texttt{<region0>}}, touching on its materials, origin, and symbolic elements. \\
    \textbf{Omni-RGPT}: The motorbike is a small dirt bike, designed for off-road racing. It is positioned on a dirt track, moving steadily across the uneven terrain. \hl{The bike is equipped with knobby tires for better traction on the loose surface.} The rider is wearing protective gear, including a helmet and a red racing suit, indicating participation in a competitive event. The background shows a dirt field with several parked vehicles and spectators, suggesting a motocross event. \hl{The bike maintains a consistent speed, navigating the bumps and turns of the course.}\\
    \rule{15.6cm}{0.5pt} \\
    \textbf{User}: Delve deeply into the description of the contents at \hlblue{\texttt{<region1>}}, focusing on the interplay of textures, colors, movements, and standout aspects. \\
    \textbf{Omni-RGPT}: The person is riding a small dirt bike, navigating through a rugged, dirt-covered terrain. They are wearing a red outfit and a helmet, indicating they are likely participating in a motocross event. \hl{The person is positioned centrally on the bike, maintaining balance as they move forward.} The background shows a dirt track with several vehicles parked, suggesting a motorsport event or competition. The person appears focused, maneuvering the bike over uneven ground, showcasing their riding skills in this challenging environment.\\
    \vspace{2mm}
    \end{tabular}
\end{minipage}
\end{tcolorbox}
}
\caption{\textbf{Qualitative visualization of detailed region-level video captioning capability.} Specifically, the sentences highlighted in orange indicate that our model effectively captures the temporal motion (or dynamics) of the targeted objects.}
\label{tab:supp_video_long_caption}
\end{table*}
\begin{table*}[t]
\centering
\resizebox{0.98\linewidth}{!}{
\begin{tcolorbox}[colback=white!100]
\centering
\begin{minipage}{\textwidth}
    \centering
    \includegraphics[width=0.95\textwidth]{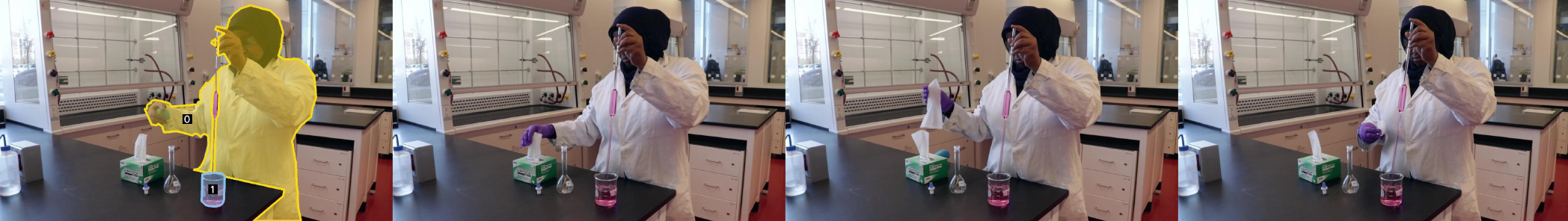} \\
    \vspace{1.5mm}
    \begin{tabular}{p{0.95\textwidth}}
    \textbf{User}: What is the setting of the scene?\\
    \textbf{Omni-RGPT}: The scene is set in a laboratory environment with a clean and organized workspace.\\
    \rule{15.6cm}{0.5pt} \\
    \textbf{User}: What objects are near \hlyellow{\texttt{<region0>}}? \\
    \textbf{Omni-RGPT}: The person is near a laboratory bench with various laboratory equipment, including a box of tissues, a beaker, and a bottle.\\
    \rule{15.6cm}{0.5pt} \\
    \textbf{User}: Could you explain what \hlyellow{\texttt{<region0>}} is doing in the scene? \\
    \textbf{Omni-RGPT}: The person is engaged in a laboratory experiment, handling a test tube filled with a pink liquid.\\
    \rule{15.6cm}{0.5pt} \\
    \textbf{User}: What is \hlyellow{\texttt{<region0>}} wearing, and what is it used for? \\
    \textbf{Omni-RGPT}: The person is wearing a white lab coat and purple gloves, which are used for protection in a laboratory setting.\\
    \rule{15.6cm}{0.5pt} \\
    \textbf{User}: What is the object \hlyellow{\texttt{<region0>}} is holding in their left hand and what is used for? \\
    \textbf{Omni-RGPT}: The person is holding a dropper in their left hand, which is used for precise liquid measurements.\\
    \rule{15.6cm}{0.5pt} \\
    \textbf{User}: What is the purpose of the object in region \hlblue{\texttt{<region1>}}? \\
    \textbf{Omni-RGPT}: The beaker is used for mixing or measuring substances.\\
    \vspace{4mm}
    \includegraphics[width=0.95\textwidth]{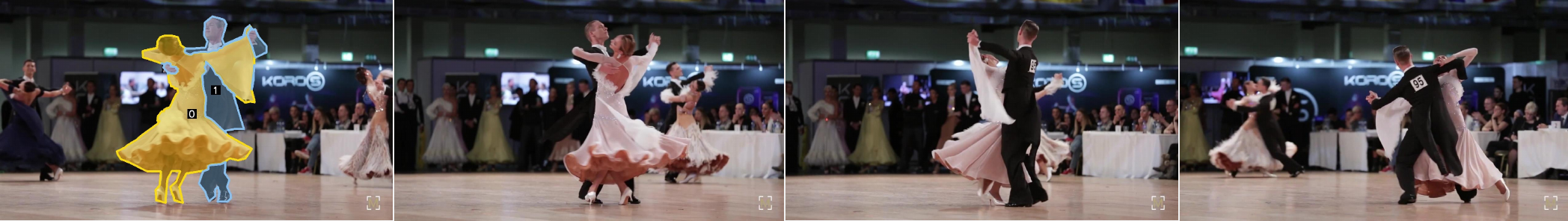} \\
    \vspace{0.3mm}
    \textbf{User}: Describe the setting in the video. \\
    \textbf{Omni-RGPT}: The setting is a ballroom with a polished wooden floor, surrounded by an audience seated at tables, creating a formal and elegant atmosphere.\\
    \rule{15.6cm}{0.5pt} \\
    \textbf{User}: Is there any interaction visible between the regions \hlyellow{\texttt{<region0>}} and \hlblue{\texttt{<region1>}}? \\
    \textbf{Omni-RGPT}: Yes, the couple is engaged in a synchronized dance routine, with the woman in a flowing dress and the man in a formal suit.\\
    \rule{15.6cm}{0.5pt} \\
    \textbf{User}: What can you say about the movement of the region \hlyellow{\texttt{<region0>}}?\\
    \textbf{Omni-RGPT}: The person is engaged in a graceful dance, moving fluidly across the dance floor.\\
    \rule{15.6cm}{0.5pt} \\
    \textbf{User}: Where is the region \hlblue{\texttt{<region1>}} located in the frame at the beginning of the video?\\
    \textbf{Omni-RGPT}: The person is positioned towards the center of the dance floor.\\
    \rule{15.6cm}{0.5pt} \\
    \textbf{User}: Where is the region \hlblue{\texttt{<region1>}} positioned in the frame at the end of the video?\\
    \textbf{Omni-RGPT}: The person is positioned towards the center-right of the frame.\\
    \end{tabular}
\end{minipage}
\end{tcolorbox}
}
\caption{\bfseries Qualitative result of region-level video QA.}
\label{tab:supp_video_qa}
\end{table*}
\begin{table}[t]
\centering
\resizebox{1.0\linewidth}{!}{
\begin{tcolorbox}[colback=white!100]
\centering
\begin{minipage}{\textwidth}
    \centering
    \includegraphics[width=6.8cm]{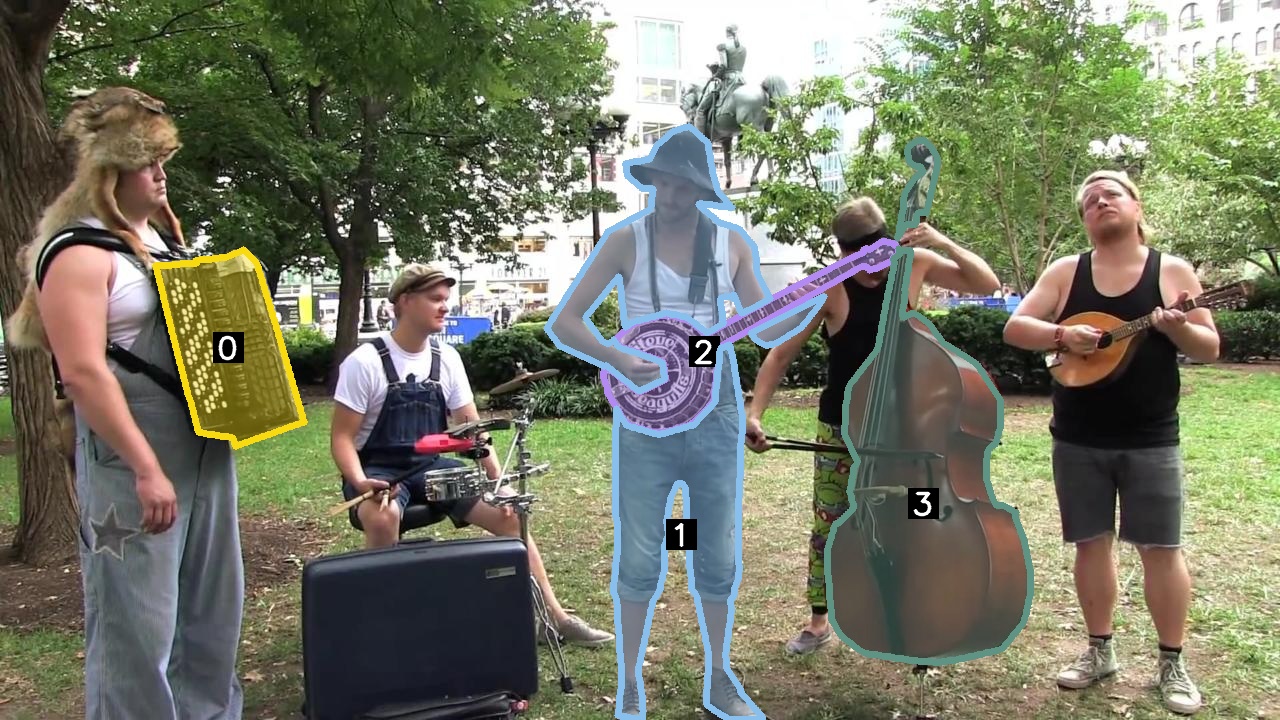}
    \begin{tabular}{p{0.95\columnwidth} c}
    \vspace{1mm}
    $\bullet$ \textbf{Brief captioning} \\
    \vspace{0.1mm}
    \textbf{User}: Can you briefly explain what you see in the region \hlyellow{\texttt{<region0>}}?\\
    \textbf{Omni-RGPT}: A black and white accordion. \\
    \rule{7.2cm}{0.5pt} \\
    \textbf{User}: What are the key features or subjects captured in this region \hlblue{\texttt{<region1>}}?\\
    \textbf{Omni-RGPT}: A man in a white tank top and blue jean shorts playing a banjo. \\
    \vspace{0.1mm}
    $\bullet$ \textbf{Detailed captioning} \\
    \vspace{0.1mm}
    \textbf{User}: Zoom in on the area indicated by \hlpurple{\texttt{<region2>}} and describe every discernible detail, from texture and color to form and function.\\
    \textbf{Omni-RGPT}: The banjo in the close-up is a large, round, and wooden instrument. It has a distinctive shape, with a curved body and a long neck. The banjo is held by a person in the group, who is likely playing it as part of the musical performance. The banjo's color is brown, which is a common color for wooden instruments. The close-up view of the banjo allows for a detailed examination of its design and craftsmanship, showcasing its unique features and the skill of the musician playing it.\\
    \rule{7.2cm}{0.5pt} \\
    \textbf{User}: Craft a thorough narrative about the piece of the image highlighted by \hlgreen{\texttt{<region3>}}, from its aesthetic qualities to its possible historical context.\\
    \textbf{Omni-RGPT}: The double bass in the close-up is a large, wooden instrument with a brown color. It has a distinctive shape, with a curved body and a long neck. The double bass is held by a person in the group, who is likely playing it as part of the musical performance. The instrument's size and style suggest that it is a traditional, acoustic double bass, which is commonly used in jazz and classical music.\\
    \end{tabular}
\end{minipage}
\end{tcolorbox}
}
\caption{\bfseries Qualitative result of region-level image captioning.}
\label{tab:supp-img-captioning}
\end{table}
\begin{table}[t]
\centering
\resizebox{1.0\linewidth}{!}{
\begin{tcolorbox}[colback=white!100]
\centering
\begin{minipage}{\textwidth}
    \centering
    \includegraphics[width=6.8cm]{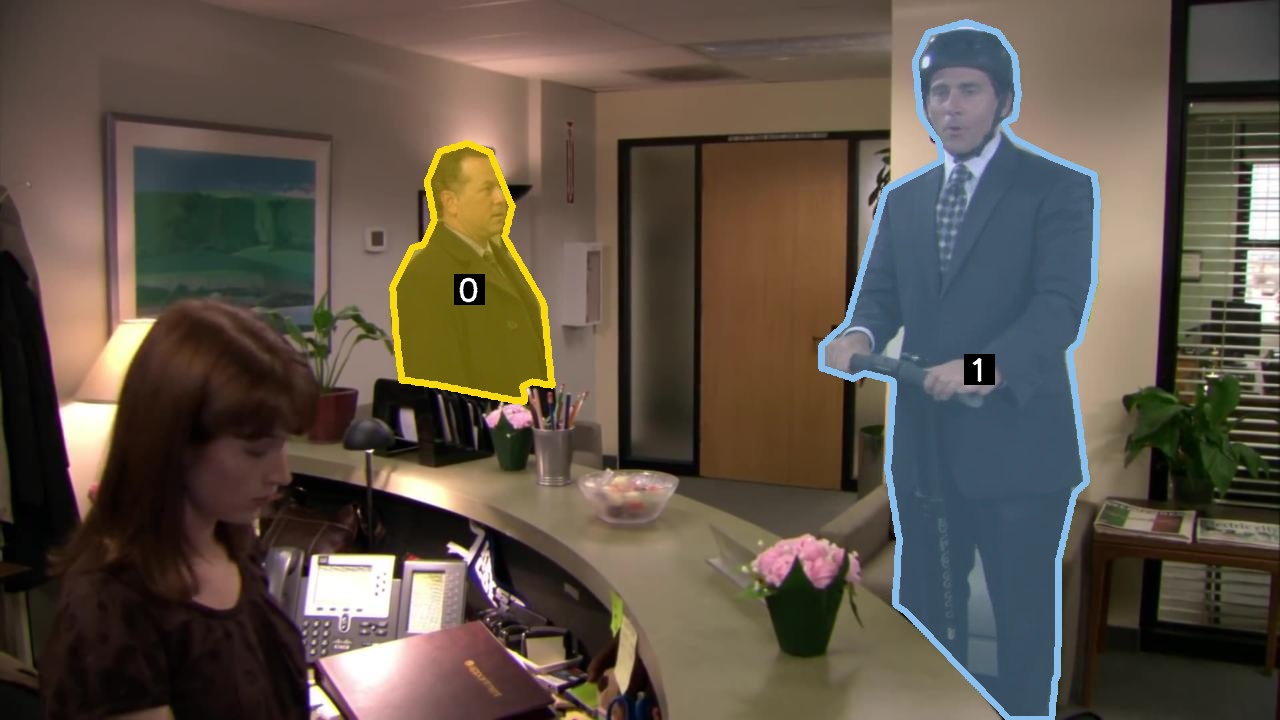}
    \begin{tabular}{p{0.95\columnwidth} c}
    \vspace{1mm}
    \textbf{User}: Why \hlyellow{\texttt{<region0>}} is looking at \hlblue{\texttt{<region1>}}? \\ 
    \textbf{Omni-RGPT}: The man in the suit is looking at the man in the helmet, \hl{possibly because he is wearing a helmet in an office setting, which is an unusual sight.} This could be a conversation starter or a point of interest for the man in the suit.\\
    \rule{7.2cm}{0.5pt} & \\
    \textbf{User}: Please explain why \hlblue{\texttt{<region1>}}'s outfit is unusual in office.\\ 
    \textbf{Omni-RGPT}: The man's outfit is unusual in the office because he is wearing a suit and a helmet. Typically, suits are associated with formal or professional attire, while helmets are more commonly worn for safety purposes, such as in construction or sports. \hl{Wearing a helmet in an office setting is not a common sight and can be considered unconventional or even humorous.} This unusual combination of attire might draw attention and spark conversations among the people in the office.\\
    \rule{7.2cm}{0.5pt} & \\
    \textbf{User}: What might be the purpose of \hlblue{\texttt{<region1>}} wearing a helmet to the office?\\
    \textbf{Omni-RGPT}: The man wearing a helmet to the office might be doing so for a variety of reasons. \hl{One possibility is that he is a cyclist or a motorcyclist} who has just arrived at work and is wearing the helmet as a safety measure. \hl{Another possibility is that he is participating in a themed event or a costume party at the office}, where wearing a helmet is part of the dress code. \hl{Alternatively, he could be wearing the helmet as a prop for a presentation or a demonstration}, where the helmet serves a specific purpose related to his job or the topic of his presentation. It is essential to consider the context and the man's role in the office to better understand the purpose behind his helmet.\\
    \end{tabular}
\end{minipage}
\end{tcolorbox}
}
\caption{\bfseries Qualitative result of region-level image QA.}
\label{tab:supp-img-qa}
\end{table}
\definecolor{gtbox}{rgb}{0.52, 0.41, 0.26}

\begin{table}[t]
\centering
\resizebox{1.0\linewidth}{!}{
\begin{tcolorbox}[colback=white!100] {
\centering

{\begin{tabular}{p{1.0\columnwidth} c}
\textbf{Input}:\\
{
\includegraphics[width=6.5cm]{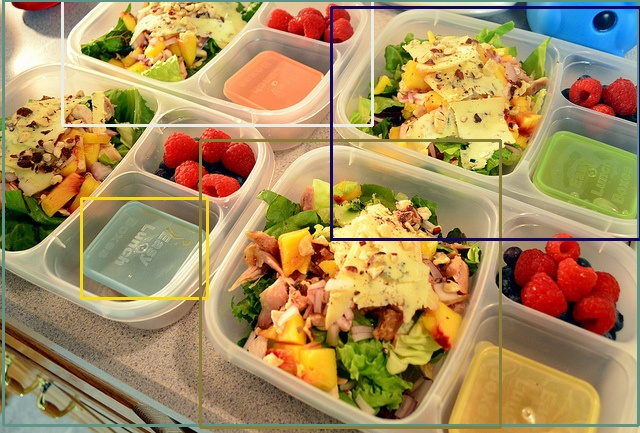}
} \\
\textbf{Caption}: \textcolor{gtbox}{the bottom right tray with a salad in it} \\ 
\rule{7.2cm}{0.5pt} & \\
\textbf{Omni-RGPT}: \\ 
\includegraphics[width=6.5cm]{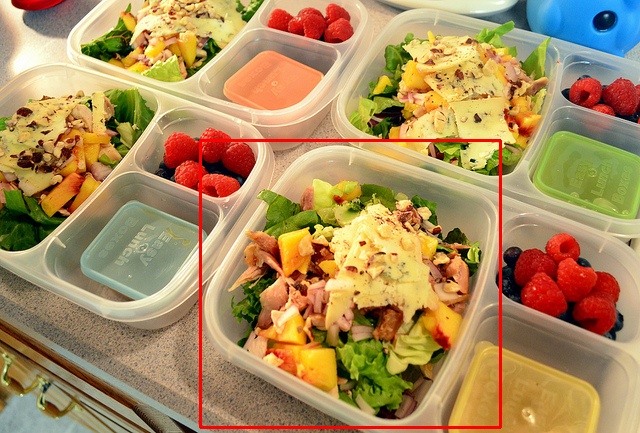}\\
\rule{7.2cm}{0.5pt} & \\
\textbf{RegionGPT}: \\ 
\includegraphics[width=6.5cm]{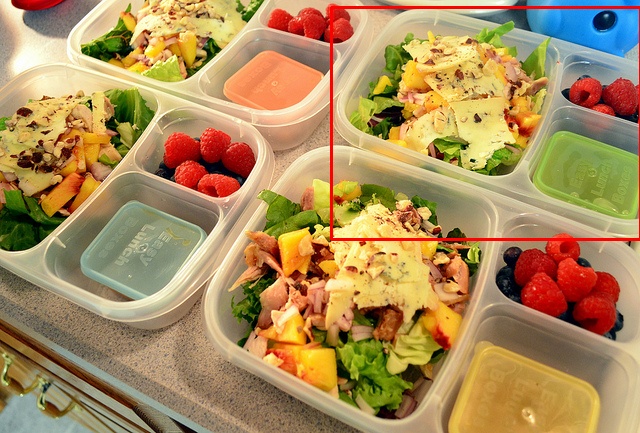}\\
\rule{7.2cm}{0.5pt} & \\
\textbf{Groma}: \\ 
\includegraphics[width=6.5cm]{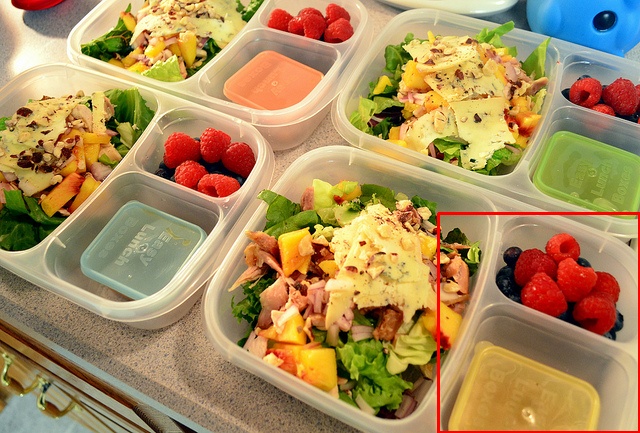}\\
\end{tabular}}}
\end{tcolorbox}
}
\caption{\bfseries Qualitative comparison of image REC.}

\label{tab:supp-img-rec-comp}
\end{table}
\definecolor{gtbox}{rgb}{0.52, 0.41, 0.26}

\begin{table}[t]
\centering
\resizebox{1.0\linewidth}{!}{
\begin{tcolorbox}[colback=white!100] {
\centering

{\begin{tabular}{p{1.0\columnwidth} c}
\textbf{Input}:\\
{
\includegraphics[width=5.85cm]{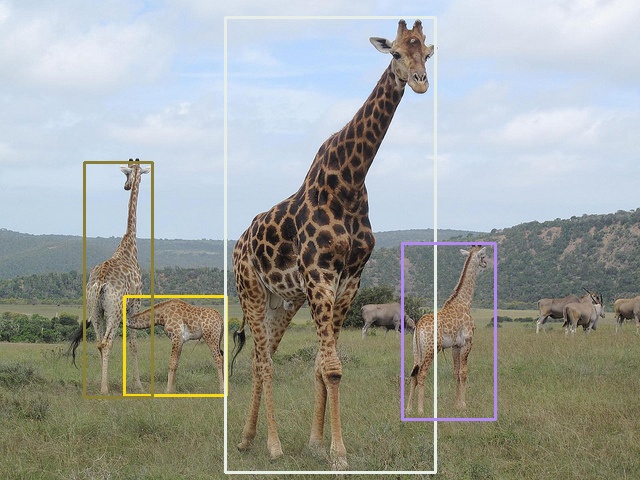}
} \\
\textbf{Caption}: \textcolor{gtbox}{a mother giraffe nursing her young} \\ 
\rule{7.2cm}{0.5pt} & \\
\textbf{Omni-RGPT}: \\ 
\includegraphics[width=5.85cm]{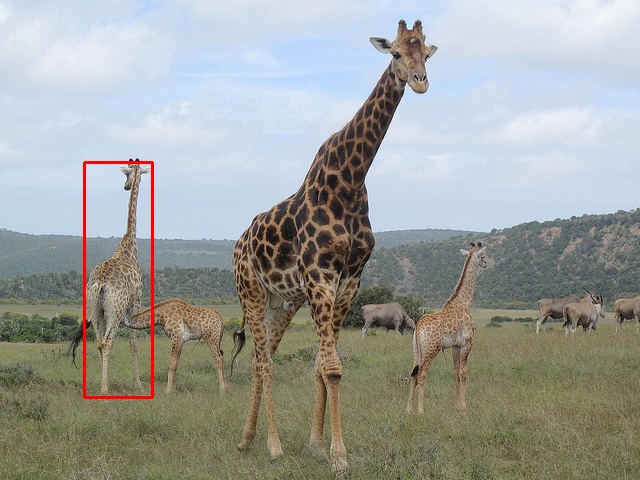}\\
\rule{7.2cm}{0.5pt} & \\
\textbf{RegionGPT}: \\ 
\includegraphics[width=5.85cm]{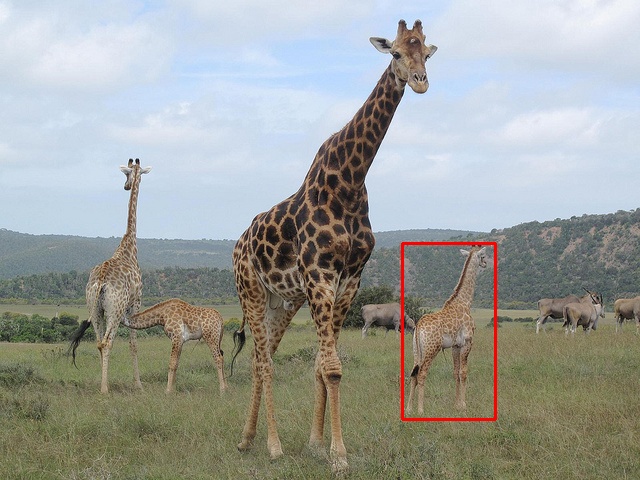}\\
\rule{7.2cm}{0.5pt} & \\
\textbf{Groma}: \\ 
\includegraphics[width=5.85cm]{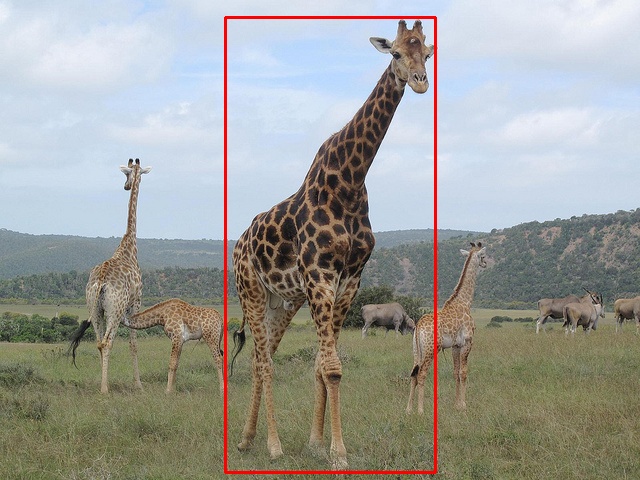}\\
\end{tabular}}}
\end{tcolorbox}
}
\caption{\bfseries Qualitative comparison of image REC.}

\label{tab:supp-img-rec-comp-2}
\end{table}
\begin{table*}[!h]
\centering
\resizebox{0.98\linewidth}{!}{
\begin{tcolorbox}[colback=white!100]
\centering
\begin{minipage}{\textwidth}
    \centering
    \vspace{2mm}
    \includegraphics[width=0.95\textwidth]{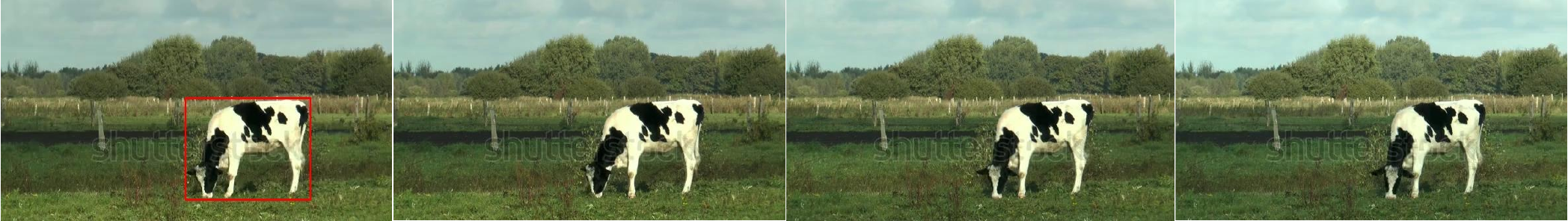} \\
    \vspace{1mm}
    \begin{tabular}{p{0.95\textwidth}}
    $\bullet$ \textbf{Original Elysium} \\
    \vspace{0.1mm}
    \textbf{User}: Can you briefly explain what you see in the \texttt{<region>}?\\
    \textbf{Ground Truth}: cattle \\
    \textbf{Omni-RGPT}: A black and white cow grazes peacefully in a serene, rural field.\\
    \rule{15.6cm}{0.5pt} \\
    $\bullet$ \textbf{Extended-Elysium} \\
    \vspace{0.1mm}
    \textbf{User}: Explain the key attributes and features of the subject found in \texttt{<region>} with in-depth detail.\\
    \textbf{Ground truth}: The cattle depicted in the series of frames is grazing steadily in a lush, green pasture. Positioned calmly in the foreground, it remains focused on eating the grass beneath. The vast field extends into the distance, bordered by a line of trees providing a natural barrier and creating a serene rural landscape. The bright, cloudy sky casts a gentle light across the meadow, highlighting the cattle's black and white patches. Throughout the frames, the cattle maintains its position, engrossed in the act of feeding.\\
    \textbf{Omni-RGPT}: The cow is positioned in the middle of a lush green field, surrounded by a backdrop of dense trees under a clear sky. The cow is grazing, with its head lowered towards the ground, indicating it is feeding on the grass. The cow's body is mostly white with distinct black patches, and it appears to be calmly moving its head as it feeds. The field is expansive, with a few scattered trees and a fence line visible in the distance, suggesting a rural setting. The cow's motion is gentle and unhurried, typical of a peaceful pastoral scene.\\
    \vspace{1mm}
    \includegraphics[width=0.95\textwidth]{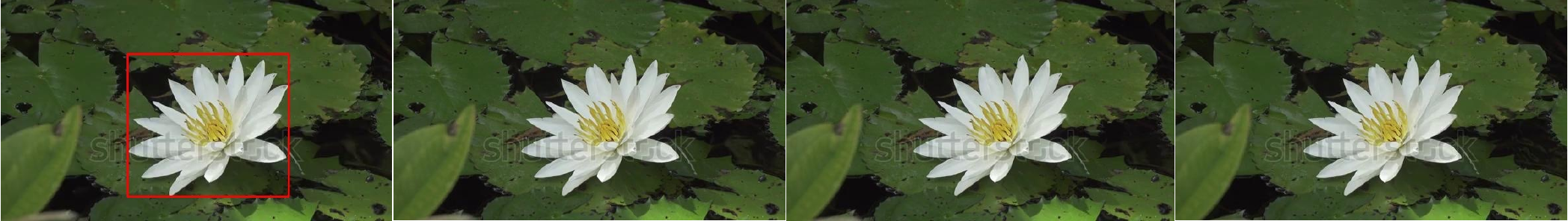} \\
    $\bullet$ \textbf{Original Elysium} \\
    \vspace{0.1mm}
    \textbf{User}: What short caption would best capture the details of \texttt{<region>}.\\
    \textbf{Ground truth}: flower \\
    \textbf{Omni-RGPT}: A white water lily with a yellow center blooms serenely on a pond surrounded by lily pads.\\
    \rule{15.6cm}{0.5pt} \\
    $\bullet$ \textbf{Extended-Elysium} \\
    \vspace{0.1mm}
    \textbf{User}: Offer a full description of the object found in \texttt{<region>}.\\
    \textbf{Ground truth}: The flower, likely a water lily, is prominently positioned atop broad, flat green leaves on the water's surface. Its pristine white petals radiate outward from a vibrant yellow center, creating a striking contrast against the leafy backdrop. In the serene water setting, the flower remains still, exhibiting a delicate and serene presence. The surrounding green leaves occasionally display small imperfections, adding a natural, untouched feel to the scene. The overall environment is tranquil, exemplifying a typical pond ecosystem where such flowers thrive.\\
    \textbf{Omni-RGPT}: The water lily is prominently positioned in the center of the frame, displaying its vibrant white petals that are slightly ruffled, suggesting a gentle breeze. The lily is surrounded by large, broad green leaves that float serenely on the water's surface, providing a lush, natural backdrop. The water lily remains stationary, its petals radiating outward, creating a symmetrical and elegant appearance. The scene is set in a tranquil pond environment, with the water reflecting the surrounding greenery, enhancing the serene and picturesque atmosphere.\\
    \end{tabular}
\end{minipage}
\end{tcolorbox}
}
\caption{\textbf{Visualized analysis of Elysium dataset.} \textbf{Top:} The original Elysium annotations are short and predominantly noun-centric.
\textbf{Bottom:} We present Extended-Elysium, refined annotations created using our dataset curation pipeline.}
\label{tab:supp_elysium}
\end{table*}
\begin{table*}[t]
\centering
\resizebox{0.98\linewidth}{!}{
\begin{tcolorbox}[colback=white!100]
\centering
\begin{minipage}{\textwidth}
    \begin{tabular}{p{0.95\textwidth}}
    $\bullet$ \textbf{Small object understanding} \\
    \vspace{0.5mm}
    \includegraphics[width=0.95\textwidth]{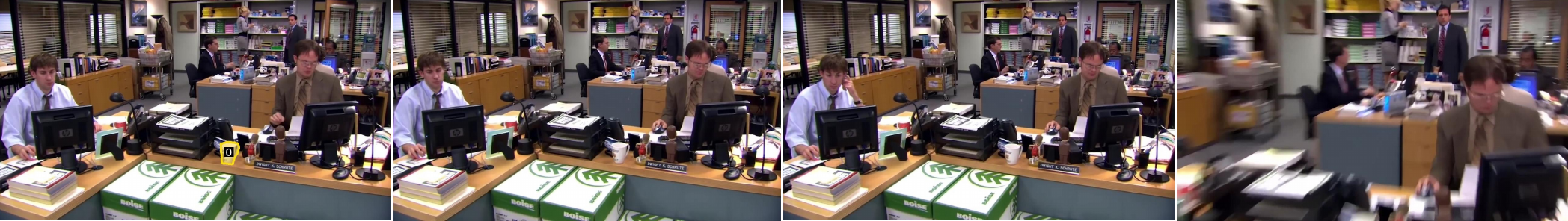} \\
    \textbf{User}: Please provide a short depiction of the content shown in \hlyellow{\texttt{<region0>}} in the video.\\
    \textbf{(\textcolor{nvgreen}{\cmark}) Omni-RGPT}: \textcolor{nvgreen}{There is a white cup} above a table in the office. \\
    \vspace{1mm}
    \includegraphics[width=0.95\textwidth]{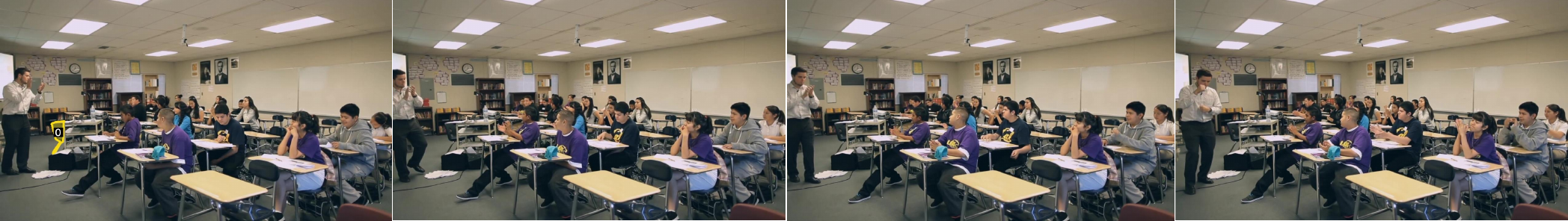} \\
    \textbf{User}: Provide a description of what is presented in the \hlyellow{\texttt{<region0>}}.\\
    \textbf{(\textcolor{red}{\xmark}) Omni-RGPT}: \textcolor{red}{A stationary backpack} is on the floor near a desk in a classroom.\\
    \vspace{2mm}
    $\bullet$ \textbf{Direction understanding in video} \\
    \vspace{0.5mm}
    \includegraphics[width=0.95\textwidth]{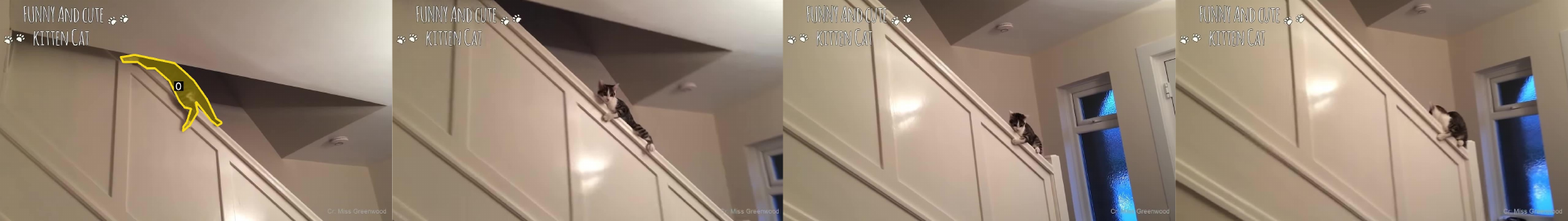} \\
    \vspace{1mm}
    \textbf{User}: Give an extensive description of the area labeled \hlyellow{\texttt{<region0>}}, paying attention to textures, colors, movements, and notable details.\\
    \textbf{(\textcolor{red}{\xmark}) Omni-RGPT}: The cat is positioned on a staircase, \textcolor{red}{appearing to be in motion as it climbs upwards}. Its body is oriented towards the top of the stairs, suggesting it is either ascending or exploring the area. The setting is indoors, with a light-colored wall and ceiling visible in the background. The cat's movement is fluid, indicating a sense of curiosity or playfulness. The scene is well-lit, with a window nearby, adding to the cozy and domestic atmosphere.\\
    \end{tabular}
\end{minipage}
\end{tcolorbox}
}
\caption{\textbf{Failure Cases}. \textbf{Top:} Omni-RGPT occasionally exhibits unstable performance on small objects. The model mistakenly identifes the black chair as a backpack in the second example.
\textbf{Bottom:} Omni-RGPT shows limited capacity for understanding the direction of objects. The model answers that the cat is climbing upward, even though it is actually moving downward.}
\label{tab:supp_video_failure}
\end{table*}
\begin{table*}[t]
\centering
\resizebox{0.98\linewidth}{!}{
\begin{tcolorbox}[colback=white!100]
\centering
\begin{minipage}{\textwidth}
    \begin{tabular}{p{0.95\textwidth}}
    \vspace{0.5mm}
    \includegraphics[width=0.95\textwidth]{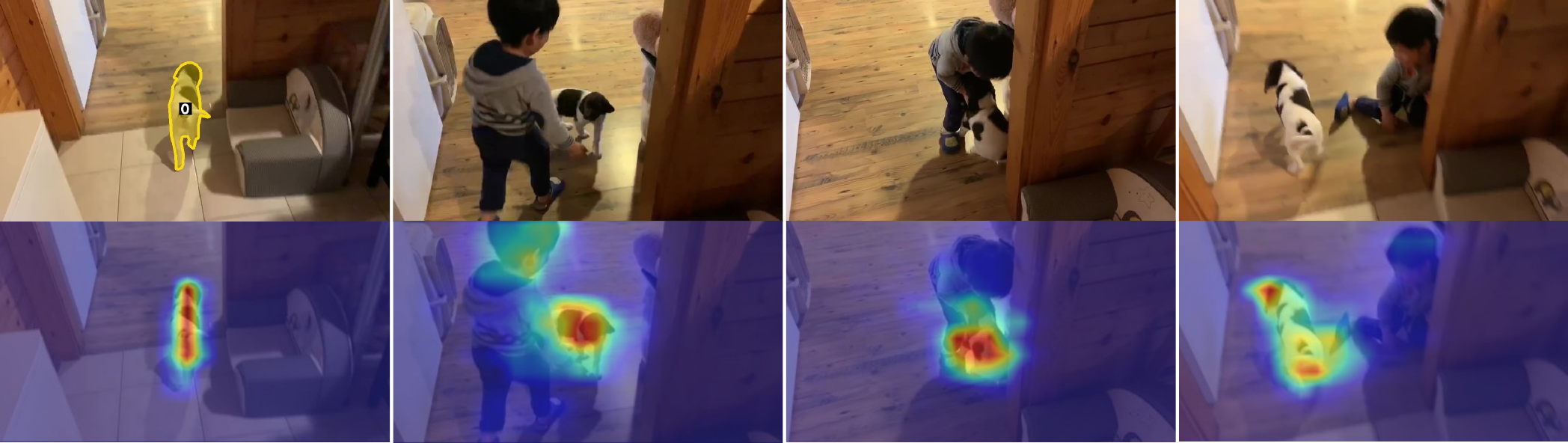} \\
    \textbf{User}: A short description of \hlyellow{\texttt{<region0>}} from the video:\\
    \textbf{Omni-RGPT}: There is a dog in white and black towards a child in gray. \\
    \vspace{1mm}
    \includegraphics[width=0.95\textwidth]{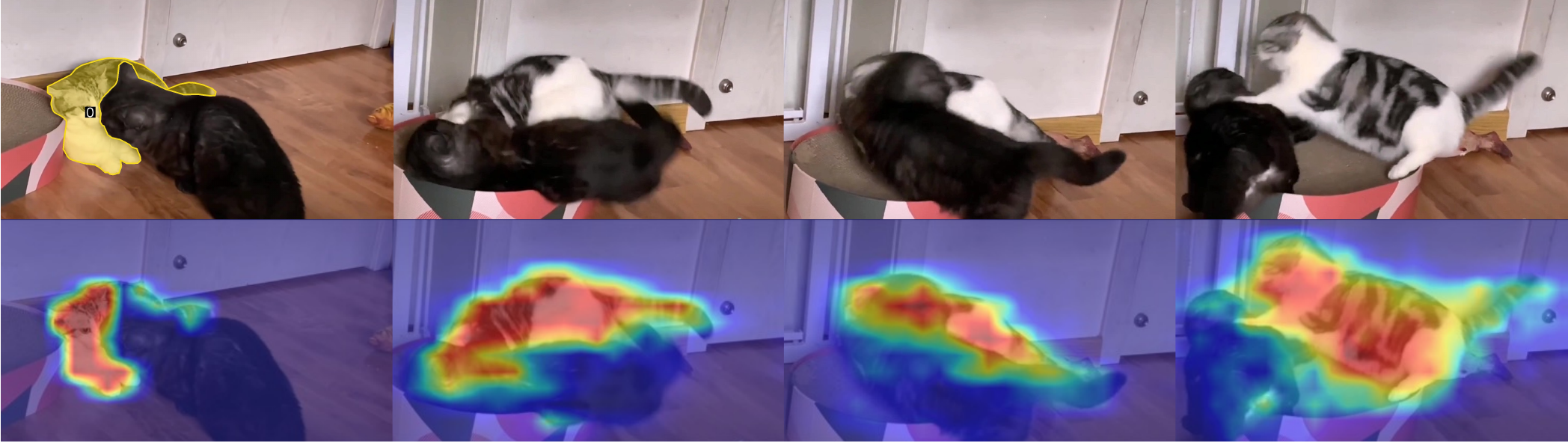} \\
    \textbf{User}: What stands out to you the most in this \hlyellow{\texttt{<region0>}}?\\
    \textbf{Omni-RGPT}: There is a cat hitting another cat on the floor.\\
    \vspace{1mm}
    \includegraphics[width=0.95\textwidth]{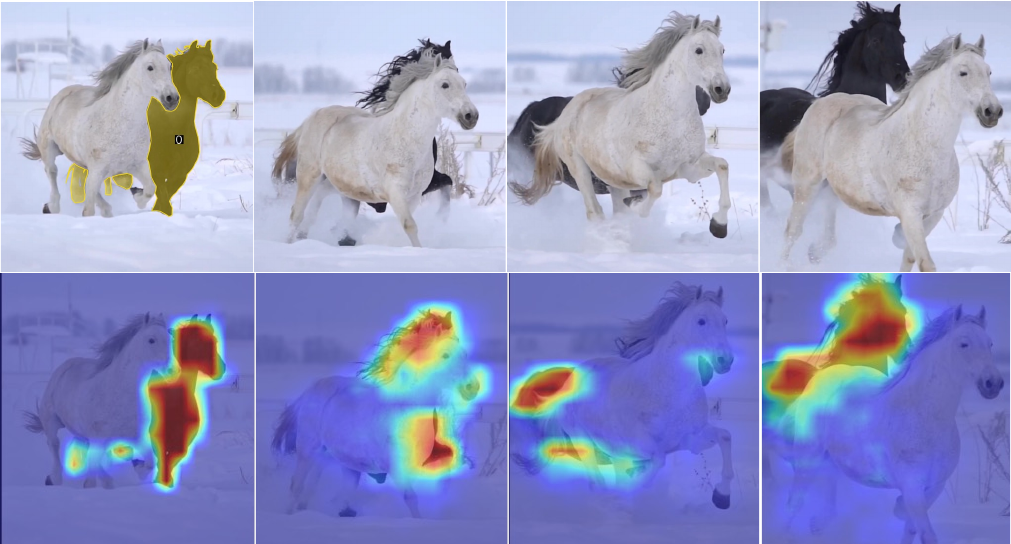} \\
    \textbf{User}: What stands out to you the most in this \hlyellow{\texttt{<region0>}}?\\
    \textbf{Omni-RGPT}: A black horse gallops energetically through a snowy landscape.\\
    \end{tabular}
\end{minipage}
\end{tcolorbox}
}
\caption{\textbf{Heatmap visualization of Temporal Region Guide Head on occluded objects}.}
\label{tab:supp_occluded_case}
\end{table*}

\begin{table*}[t]
\centering
\resizebox{0.98\linewidth}{!}{
\begin{tcolorbox}
\centering
\begin{minipage}{\textwidth}
    \centering
    \includegraphics[width=0.95\textwidth]{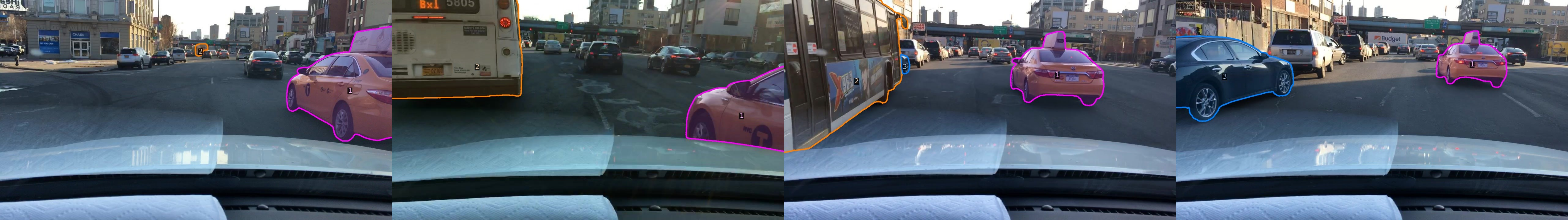} \\[0.5em]
    \vspace{1mm}
    \begin{tabular}{p{0.95\textwidth}}
    \textbf{User}: Generate a long, detailed caption for the segment of the video at \hlmaskpink{\texttt{<region1>}}, covering aspects such as its origin, material, and any symbolic meaning. \\
    \textbf{Assistant}: The car is a yellow taxi positioned on the right side of the road. It stays largely consistent in position, seemingly moving at a steady pace parallel to the dashcam car. The taxi's presence remains constant, indicating it's traveling in the same general direction as the surrounding traffic. The background suggests an urban environment, with multiple lanes and buildings visible, typical of a city street. \\
    \rule{15.6cm}{0.5pt} \\
    \textbf{User}: Expand upon the visual description of what is contained within the marked area \hlmaskorange{\texttt{<region2>}}, covering its appearance, size, and stylistic features. \\
    \textbf{Assistant}: The bus is located ahead of the taxi and further up the road. Initially, it appears small due to its distance, but gradually becomes larger and more detailed as the dashcam car advances down the road. The bus is traveling in the same direction as the rest of the traffic and navigating through the urban setting. Close proximity to other vehicles and buildings in the scene are evident as they approach a more congested area. \\
    \rule{15.6cm}{0.5pt} \\
    \textbf{User}: Explain the key attributes and features of the subject found in \hlmaskblue{\texttt{<region3>}} with in-depth detail. \\
    \textbf{Assistant}: The car is a black sedan situated to the left of the bus, becoming visible later in the sequence of frames. Initially, it is further behind but gradually moves up to parallel with the dashcam car. The black sedan is traveling in the same direction as the taxi and bus within the lanes. It doesn't exhibit any erratic movements, maintaining a steady course. The scene's background remains consistently urban, with the overhead signage and multiple vehicles indicating a busy city road.\\
    \rule{15.6cm}{0.5pt} \\
    \textbf{User}: How would you briefly summarize the key points in \hlmaskpink{\texttt{<region1>}}? \\
    \textbf{Assistant}: A yellow taxi is traveling steadily alongside the dashcam car on a city street. \\
    \rule{15.6cm}{0.5pt} \\
    \textbf{User}: Describe what's highlighted within \hlmaskorange{\texttt{<region2>}}. \\
    \textbf{Assistant}: A bus is ahead of the taxi and grows larger as the dashcam car approaches in urban traffic. \\
    \rule{15.6cm}{0.5pt} \\
    \textbf{User}: Write a short description for the region \hlmaskblue{\texttt{<region3>}}. \\
    \textbf{Assistant}: A black sedan becomes parallel with the dashcam car, moving steadily in city traffic. \\
    \rule{15.6cm}{0.5pt} \\
    \textbf{User}: What color is the taxi in region \hlmaskpink{\texttt{<region1>}}? \\
    \textbf{Assistant}: The taxi is yellow.\\
    \rule{15.6cm}{0.5pt} \\
    \textbf{User}: Where is the taxi positioned relative to the road? \\
    \textbf{Assistant}: The taxi is positioned on the right side of the road. \\
    \rule{15.6cm}{0.5pt} \\
    \textbf{User}: Is the yellow taxi moving erratically? \\
    \textbf{Assistant}: No, the taxi is moving at a steady pace parallel to the dashcam car. \\
    \rule{15.6cm}{0.5pt} \\
    \textbf{User}: Where is the bus located in relation to the taxi? \\
    \textbf{Assistant}: The bus is located ahead of the taxi and further up the road. \\
    \rule{15.6cm}{0.5pt} \\
    \textbf{User}: When does the black sedan in region \hlmaskblue{\texttt{<region3>}} become visible? \\
    \textbf{Assistant}: The black sedan becomes visible later in the sequence of frames. \\
    \end{tabular}
\end{minipage}
\end{tcolorbox}
}
\caption{\textbf{Visualized instruction-following sample and the video in \datasetabbr.} The object masks are overlayed onto the video frames. The video is from the BDD100k dataset~\cite{BDD100k}.}
\label{tab:data_sample_1}
\end{table*}
\begin{table*}[t]
\centering
\resizebox{0.98\linewidth}{!}{
\begin{tcolorbox}
\centering
\begin{minipage}{\textwidth}
    \centering
    \includegraphics[width=0.95\textwidth]{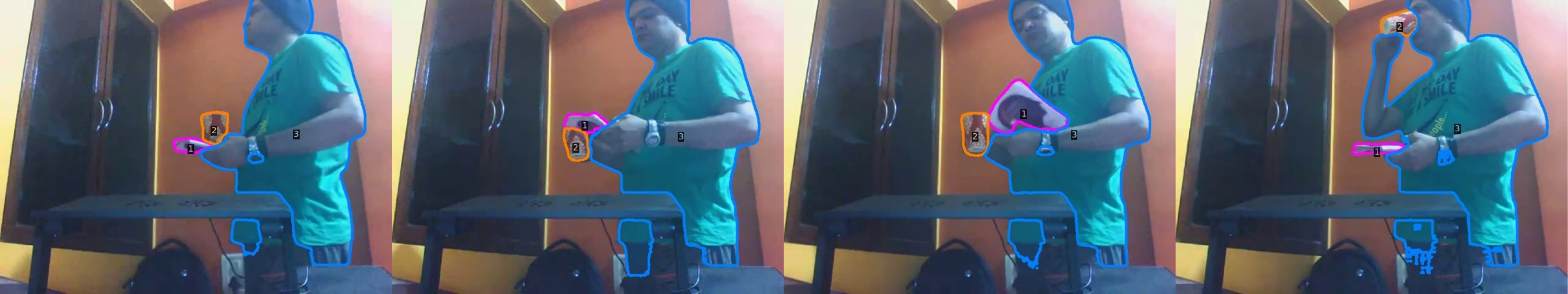} \\[0.5em]
    \vspace{1mm}
    \begin{tabular}{p{0.95\textwidth}}
    \textbf{User}: Elaborate on the artwork shown in the region indicated by \hlmaskpink{\texttt{<region0>}}, mentioning its color, appearance, size, style, and any standout features. \\
    \textbf{Assistant}: The book is held in the person's left hand. It appears to be a soft-cover book, with the pages slightly visible as they move and adjust their grip on it. The person keeps the book close to their torso while standing near a table in a room with an orange wall and a wooden cabinet in the background. As they maneuver, the book remains prominently in hand, indicating it could be light and easy to handle. \\
    \rule{15.6cm}{0.5pt} \\
    \textbf{User}: Provide a detailed account of the object located at \hlmaskorange{\texttt{<region1>}} in the video. \\
    \textbf{Assistant}: The object, looking like a cup or bottle, is initially near the center of the person's torso and gradually moves toward their mouth in the latter frames. This action suggests it might contain a beverage as they appear to be preparing to drink from it. The cup has a distinct rim, which contrasts against the orange wall, flanked by the cabinet on the left side of the frame. \\
    \rule{15.6cm}{0.5pt} \\
    \textbf{User}: Describe comprehensively the object at \hlmaskblue{\texttt{<region2>}}, including its physical traits, style, actions, and visible characteristics. \\
    \textbf{Assistant}: The person is standing, wearing a green T-shirt and a beanie. They are constantly adjusting their position while holding a book in one hand and a cup or bottle in the other, showing a casual and relaxed demeanor. The scene is set in a room with a bold orange paint and a wooden cabinet with reflective glass, indicating a cozy indoor environment. \\
    \rule{15.6cm}{0.5pt} \\
    \textbf{User}: Write a short description for the region \hlmaskorange{\texttt{<region1>}}. \\
    \textbf{Assistant}: A cup or bottle moves towards the person's mouth, suggesting a drink. \\
    \rule{15.6cm}{0.5pt} \\
    \textbf{User}: How would you briefly summarize the key points in \hlmaskblue{\texttt{<region2>}}? \\
    \textbf{Assistant}: A person in a green T-shirt holds a book and cup in a cozy room. \\
    \rule{15.6cm}{0.5pt} \\
    \textbf{User}: Describe the action being performed with the object in region \hlmaskorange{\texttt{<region1>}}. \\
    \textbf{Assistant}: The object, which looks like a cup or bottle, is gradually moving toward the person's mouth, suggesting they are preparing to drink from it. \\
    \rule{15.6cm}{0.5pt} \\
    \textbf{User}: What is the clothing color of the person in region \hlmaskblue{\texttt{<region2>}}? \\
    \textbf{Assistant}: The person is wearing a green T-shirt. \\
    \rule{15.6cm}{0.5pt} \\
    \textbf{User}: Is there any motion involving the book in region\hlmaskpink{ \texttt{<region0>}}? \\
    \textbf{Assistant}: Yes, the book slightly moves as the person adjusts their grip on it. \\
    \rule{15.6cm}{0.5pt} \\
    \textbf{User}: In what hand is the object in region \hlmaskorange{\texttt{<region1>}}? \\
    \textbf{Assistant}: The cup or bottle is in one hand, likely not the same where the book is held. \\
    \rule{15.6cm}{0.5pt} \\
    \textbf{User}: What is the backdrop environment in the entire video? \\
    \textbf{Assistant}: The environment is a room with an orange wall and a wooden cabinet with reflective glass, which gives a cozy indoor feel. \\
    \end{tabular}
\end{minipage}
\end{tcolorbox}
}
\caption{\textbf{Visualized instruction-following sample and the video in \datasetabbr.} The object masks are overlayed onto the video frames. The video is from the Action Genome dataset~\cite{ActionGenome}.}
\label{tab:data_sample_2}
\end{table*}
\begin{table*}[tb]
\centering
\begin{minipage}{2.0\columnwidth}
\begin{tcolorbox}
You are an intelligent video captioner designed to generate the description of objects. Your task is to generate detailed descriptions of the objects labeled with a bright numeric ID at the center of video frames. 
\newline \newline
\#\#INSTRUCTIONS: 

-  The description should start with the object noun. DO NOT add any irrelevant content to the answer. Each object should be described in around 90 words, but DO NOT include information that you are unaware of. 

- INCLUDE the following information in the object descriptions: (1) temporal (motion, action), (2) contextual (spatial position) information of objects, and (3) global (background, scene) information. If you cannot see the object clearly, output the following \verb|<|object id\verb|>|: None. 

- MUST NOT include the outline color information in the description. Use bracket [] to refer to the object IDs in the descriptions if needed.
\newline \newline
These are video frames. Generate descriptions of each labeled object, \verb|[|\texttt{<id1>}\verb|]|: \texttt{<noun1>}, \verb|[|\texttt{<id2>}\verb|]|: \texttt{<noun2>}..., .
\end{tcolorbox}
\end{minipage}
\caption{\textbf{Input text prompts to GPT4o for the detailed region-level captioning.} We input the above prompts and SoM-processed videos to obtain region-level captions. The placeholders \texttt{<idX>} and \texttt{<nounX>} are replaced with $X$-th object indices marked in the video and the corresponding object's ground-truth noun, respectively. }
\label{tab:gpt4o_prompt}
\end{table*}
\begin{table*}[t!]
\centering
\begin{minipage}{2.0\columnwidth}
\begin{tcolorbox}
Your task is to generate multiple questions comprehensively from given object descriptions in JSON format.
The question should ask about the object from the given descriptions. The questions comprehensively ask all the contents in the object descriptions. DO NOT generate a question whose answer will be no. Respond only with valid JSON. Do not write an introduction or summary.
\newline 

Description: \texttt{<ic sample1 caption>}  \\
Question: \{``questions'': \verb|[| \texttt{<ic sample1 q1>},  \texttt{<ic sample1 q2>}, ..., \verb|]| \} \\

Description: \texttt{<ic sample2 caption>}  \\
Question: \{``questions'': \verb|[| \texttt{<ic sample2 q1>},  \texttt{<ic sample2 q2>}, ..., \verb|]| \} \\

Description: \texttt{<caption>}  \\
Question: 
\end{tcolorbox}
\end{minipage}
\caption{\textbf{The text prompts to decompose the object-level to multiple fact-checking questions.} To enhance the output quality, we manually create in-context pairs (input object description and output questions).  The placeholder \texttt{<ic sampleX caption>} is replaced with the object descriptions for the $X$-th in-context samples, and \texttt{<ic sampleX qY>} is replaced with the manually annotated $Y$th question of $X$th in-context sample. The target object caption is injected to the placeholder \texttt{<caption>}.}
\label{tab:decomp_prompt}
\end{table*}
\begin{table*}[tb]
\centering
\begin{minipage}{2.0\columnwidth}
\begin{tcolorbox}
\texttt{<Q>} Answer from the following options: (A): Definitely Yes, (B): Possibly Yes, (C): Possibly No, (D): Definitely No, (E): Not sure. Do not write an introduction or summary. Only respond the letter.
\end{tcolorbox}
\end{minipage}
\caption{\textbf{The text prompt to input to the MLLM to verify the contents in the captions.} The placeholder \texttt{<Q>} is replaced with the input questions generated during the previous stage.}
\label{tab:fact_checking_prompt}
\end{table*}
\begin{table*}[tb]
\centering
\begin{minipage}{2.0\columnwidth}
\begin{tcolorbox}
Your task is to generate a refined description of the objects based on the given questions about the objects in a list format. The answer is always no, so remove that information from the object description and re-generate the refined description. Do not change the irrelevant content as much as possible. Do not write an introduction or summary, and only write the description in JSON format.
\newline
======= \\
Description:  \texttt{<ic sample original caption>}  \\
Questions:  \verb|[| \texttt{<ic sample q1>},  \texttt{<ic sample q2>}, ..., \verb|]| \\
Refined Description: \texttt{<ic sample output caption>}  \\

====== \\
Description:  \texttt{<original caption>}  \\
Questions:  \verb|[| \texttt{<q1>},  \texttt{<q2>}, ..., \verb|]| \\
Refined Description:
\end{tcolorbox}
\end{minipage}
\caption{\textbf{The text prompt to refine the region-level captions.} We collect the questions whose answers are ``D'' in ~\cref{tab:fact_checking_prompt} and prompt the LLM to remove the contents in the questions from the original caption. One in-context sample is provided to improve the output quality. The placeholders \texttt{<ic sample original caption>}, \texttt{<ic sample qX>}, \texttt{<ic sample output caption>} are replaced with the object-level caption, the $X$-th question which includes the non-verified content, and the manually annotated refined caption, respectively. The target caption and non-verified questions are injected to \texttt{<original caption>} and \texttt{<qX>} placeholders, respectively.}
\label{tab:regenerate_cap}
\end{table*}
\begin{table*}[tb]
\centering
\begin{minipage}{2.0\columnwidth}
\begin{tcolorbox}
You are a helpful assistant in summarizing the long description of the object into a single sentence or a short description. Output the JSON format, for example, {\verb|[<|object ID\verb|>]|: \verb|<|description\verb|>|, ...} \\ 
\#\#Instructions \\ 
- DO NOT include any contents that have not appeared in the long descriptions. \\ 
- The short description should briefly cover the abstracted contents of the long description. \\ 
- The short description should be a single sentence, and the number of words had better be less than 20. \\  \\
The long descriptions of objects (\texttt{<id1>}, \texttt{<id2>}, ...) are as follows: \verb|[|\texttt{<id1>}\verb|]|: \texttt{<description1>}, \verb|[|\texttt{<id2>}\verb|]|: \texttt{<description2>}, ...
\end{tcolorbox}
\end{minipage}
\caption{\textbf{The text prompt to create a summary of the objects' descriptions.} The placeholders \texttt{<idX>} and \texttt{<descriptionX>} are replaced with $X$-th object region indices and the corresponding detailed caption.}
\label{tab:summary_prompt}
\end{table*}
\begin{table*}[tb]
\centering
\begin{minipage}{2.0\columnwidth}
\begin{tcolorbox}
You are a helpful assistant in creating diverse multi-round conversation question-and-answer pairs from the given object descriptions. Your task is to design a conversation between you and a person asking about objects. The answers should be in a tone that a visual AI assistant is watching the video and answering the question. Output the JSON format with key 'Q' and value 'A', for example, \verb|{|'QA\verb|_|pairs': \verb|[{|'Q': 'What is the object next to region \verb|[0]|...', 'A': 'It is...'\verb|}|,\verb|{|'Q': 'Why this object is...'\verb|}| ...\verb|]}|. \\

Include diverse questions asking about the visual content of the video, including the object types, temporal information, object motions, actions, textures, spatial positions, etc. Only include questions that have definite answers: \\ 
(1) one can see the content in the video that the question asks about and can answer confidently; \\ 
(2) One can confidently determine from the video that it is not in the video. \\ 
Do not ask any questions that cannot be answered confidently. \\
Please follow the below instructions. \\
\#\# Instructions \\
- Do not add additional information to the output answer from the input object descriptions. DO NOT ask about uncertain details. \\
- If you refer to a certain object from the list, USE region indices (\texttt{<id1>}, \texttt{<id2>}, ...) in the questions. DO NOT include the region indices in the answer. MUST generate several different QAs. \\ \\ 
The object descriptions for \texttt{<id1>}, \texttt{<id2>}, ... are as below: \verb|[|\texttt{<id1>}\verb|]|: \texttt{<description1>}, \verb|[|\texttt{<id2>}\verb|]|: \texttt{<description2>}, ...
\end{tcolorbox}
\end{minipage}
\caption{\textbf{The text prompt to generate region-level conversation samples from object-level detailed captions.} The placeholders \texttt{<idX>} and \texttt{<descriptionX>} are replaced with $X$-th object region indices and the corresponding detailed caption.}
\label{tab:conv_prompt}
\end{table*}
\begin{table*}[tb]
\centering
\begin{minipage}{2.0\columnwidth}
\begin{tcolorbox}
\scalebox{0.68}{\vbox{%
\begin{itemize}
\item Provide a brief caption for the area indicated by \texttt{<region>}.
\item Describe in a short phrase the content within the bounds of \texttt{<region>}.
\item How would you succinctly caption the region highlighted by \texttt{<region>}?
\item Summarize the scene or object present in the section marked by \texttt{<region>}.
\item Can you give a concise description of what’s depicted in \texttt{<region>}?
\item Draft a short title for the video content enclosed by \texttt{<region>}.
\item What brief caption would best describe the visual within \texttt{<region>}?
\item Offer a succinct interpretation of the area pointed out by \texttt{<region>}.
\item If you were to provide a short tagline for the content at \texttt{<region>}, what would it be?
\item Give a one-liner description of the region demarcated by \texttt{<region>}.
\item How would you encapsulate the essence of the segment labeled \texttt{<region>} in a few words?
\item Characterize the content of the video portion specified by \texttt{<region>} briefly.
\item Craft a mini headline for the visual element spotlighted by \texttt{<region>}.
\item In a few words, how would you narrate the content found within \texttt{<region>}?
\item Pen down a concise caption for the video section delineated by \texttt{<region>}.
\item A short caption of region \texttt{<region>}: 
\item A short description of region \texttt{<region>}: 
\item A photo containing the region \texttt{<region>}: 
\item A region \texttt{<region>} that shows 
\item Write a short description for the region \texttt{<region>} 
\item Write a description for the region \texttt{<region>} 
\item Provide a description of what is presented in the region \texttt{<region>}.
\item Briefly describe the content of the region \texttt{<region>}.
\item Can you briefly explain what you see in the region \texttt{<region>}?
\item Could you use a few words to describe what you perceive in the region \texttt{<region>}?
\item Please provide a short depiction of the region \texttt{<region>}.
\item Using language, provide a short account of the region \texttt{<region>}.
\item Use a few words to illustrate what is happening in the region \texttt{<region>}.
\item Provide an overview of what you see in the region \texttt{<region>}.
\item Can you break down the main elements present in this region \texttt{<region>}?
\item What are the key features or subjects captured in this region \texttt{<region>}?
\item Summarize the primary components of this region \texttt{<region>}.
\item Walk me through the different aspects of this region \texttt{<region>}.
\item Highlight the main points of interest in this region \texttt{<region>}.
\item What stands out to you the most in this region \texttt{<region>}?
\item If you were to give a brief overview of this region \texttt{<region>}, what would you mention?
\item List the primary objects or subjects you identify in this region \texttt{<region>}.
\item Describe the first few things that catch your attention in this region \texttt{<region>}.
\item How would you introduce this region \texttt{<region>} to someone who hasn’t seen it?
\item What are the defining characteristics of this region \texttt{<region>}?
\item Give a concise description of the main content in this region \texttt{<region>}.
\item If you were to caption this region \texttt{<region>}, what might you say?
\item Describe the scene or setting depicted in this region \texttt{<region>}.
\item Provide a concise label for the highlighted \texttt{<region>}.
\item Offer a short depiction of what’s enclosed within \texttt{<region>}.
\item How would you briefly label the segment outlined by \texttt{<region>}?
\item Summarize the contents of the space marked by \texttt{<region>}.
\item Can you give a brief explanation of what’s featured in \texttt{<region>}?
\item Suggest a quick headline for the area depicted within \texttt{<region>}.
\item What short caption would best capture the details of \texttt{<region>}?
\item Provide a concise overview of the highlighted portion of \texttt{<region>}.
\item If you were to summarize the content in \texttt{<region>}, what would it be?
\item Offer a one-line explanation of what is shown in \texttt{<region>}.
\item How would you briefly summarize the key points in \texttt{<region>}?
\item Characterize the details of the video enclosed within \texttt{<region>} in a few words.
\item Draft a small title for the area spotlighted in \texttt{<region>}.
\item How would you narrate the content observed in \texttt{<region>} concisely?
\item Write a brief caption for the section of the video enclosed by \texttt{<region>}.
\item Create a short label for the visible content in \texttt{<region>}.
\item Offer a quick description of the content shown in \texttt{<region>}.
\item A snapshot of what \texttt{<region>} includes.
\item Describe what’s highlighted within \texttt{<region>}.
\item Provide a quick summary of the contents present in \texttt{<region>}.
\item What’s the best way to describe the content within \texttt{<region>} briefly?
\item What’s shown in \texttt{<region>}?Please describe.
\item How would you quickly explain what is featured in \texttt{<region>}?
\item Provide a short summary of what’s visible in \texttt{<region>}.
\item Could you break down what you see in \texttt{<region>}?
\item Can you briefly outline the main elements in \texttt{<region>}?
\item Give a brief interpretation of what stands out in \texttt{<region>}.
\item What are the key points of focus in this section \texttt{<region>}?
\item Provide a concise rundown of what is visible in \texttt{<region>}.
\item Describe the notable objects or subjects you can see in \texttt{<region>}.
\item How would you introduce the details shown in \texttt{<region>} to someone new?
\item List the main features or subjects captured in \texttt{<region>}.
\item Summarize the primary components present in \texttt{<region>}.
\item What catches your attention first in \texttt{<region>}?
\end{itemize}
}}
\end{tcolorbox}
\end{minipage}
\vspace{-3mm}
\caption{\textbf{The list of instructions for brief region description.}}
\label{tab:brief_caption_templates}
\end{table*}
\begin{table*}[tb]
\centering
\begin{minipage}{2.0\columnwidth}
\begin{tcolorbox}
\scalebox{0.9}{\vbox{%
\begin{itemize}
\item Describe in detail the object located at \texttt{<region>} in the video, including its appearance, style, actions, and any visible details.
\item Provide a comprehensive description of the area marked by \texttt{<region>}, focusing on textures, colors, motions, and any notable features.
\item Elaborate on the artwork shown in the region indicated by \texttt{<region>}, mentioning its color, appearance, size, style, and any standout features.
\item Give a detailed analysis of the scene within the boundary of \texttt{<region>}, touching upon its components, ambiance, and any thematic expressions.
\item Craft a thorough narrative about the piece of the video highlighted by \texttt{<region>}, from its aesthetic qualities to its possible historical context.
\item Explain in depth the characteristics and attributes of the subject found in the segment tagged with \texttt{<region>}.
\item Generate a long, detailed caption for the segment of the video at \texttt{<region>}, covering aspects such as its origin, material, and any symbolic meaning.
\item Paint a vivid picture with words about the region at \texttt{<region>}, diving into the intricacies and nuances present in the area.
\item Zoom in on the area indicated by \texttt{<region>} and describe every discernible detail, from texture and color to form and function.
\item Offer an expanded description of the contents within the area marked by \texttt{<region>}, encompassing its color, appearance, size, style, and any remarkable features.
\item Provide a detailed account of the object located at \texttt{<region>} in the video.
\item Offer a full description of the object found in \texttt{<region>}.
\item Describe comprehensively the object at \texttt{<region>}, including its physical traits, style, actions, and visible characteristics.
\item Give an extensive description of the area labeled \texttt{<region>}, paying attention to textures, colors, movements, and notable details.
\item Elaborate thoroughly on the artwork situated in the area indicated by \texttt{<region>}, noting its color, size, style, and key features.
\item Analyze the scene within the confines of \texttt{<region>} in detail, considering its components, atmosphere, and any thematic elements.
\item Develop a rich narrative for the part of the video highlighted in \texttt{<region>}, exploring its aesthetic aspects and potential historical significance.
\item Explain the key attributes and features of the subject found in \texttt{<region>} with in-depth detail.
\item Compose an extended caption for the video segment located at \texttt{<region>}, focusing on its origin, materials, and possible symbolic meaning.
\item Describe vividly the region marked \texttt{<region>}, exploring the intricacies and nuances present in the area.
\item Focus on the area highlighted by \texttt{<region>} and provide a meticulous description, covering its texture, color, shape, and function.
\item Expand on the description of what’s present in the area of \texttt{<region>}, addressing its colors, appearance, dimensions, style, and distinctive qualities.
\item Give a precise and thorough breakdown of the object located at \texttt{<region>}, considering its visual and functional features.
\item Delve deeply into the description of the contents at \texttt{<region>}, focusing on the interplay of textures, colors, movements, and standout aspects.
\item Explore the artwork or object featured in \texttt{<region>} with a focus on its color, style, and size.
\item Offer a detailed breakdown of the scene in \texttt{<region>}, discussing its components, mood, and any visible themes.
\item Provide a rich account of the aesthetic qualities and potential historical or cultural relevance of the area highlighted by \texttt{<region>}.
\item Explain the visible details of the subject in \texttt{<region>}, including its appearance, movements, and any distinguishing traits.
\item Create a thorough, descriptive caption for the portion of the video located in \texttt{<region>}, touching on its materials, origin, and symbolic elements.
\item Use vivid language to describe the region marked by \texttt{<region>}, diving into the subtle details and unique aspects of the area.
\item Take a close look at \texttt{<region>} and describe in detail the features you observe, including texture, color, form, and function.
\item Expand upon the visual description of what is contained within the marked area \texttt{<region>}, covering its appearance, size, and stylistic features.
\end{itemize}
}}
\end{tcolorbox}
\end{minipage}
\vspace{-3mm}
\caption{\textbf{The list of instructions for detailed region description.}}
\label{tab:detailed_caption_templates}
\end{table*}

\end{document}